%% file: 0_main.tex
\documentclass[runningheads]{llncs}
\usepackage{graphicx}

\usepackage{tikz}
\usepackage{comment}
\usepackage{amsmath,amssymb} 
\usepackage{color}

\usepackage{booktabs}
\DeclareMathOperator*{\argmax}{argmax}
\usepackage{multirow}
\usepackage{soul}
\usepackage{nicefrac}
\usepackage{diagbox}
\usepackage{float}
\usepackage{authblk}

\usepackage{array}
\usepackage{enumitem}
\newcolumntype{?}{!{\vrule width 1pt}}

\usepackage{amssymb}
\usepackage{pifont}
\usepackage[accsupp]{axessibility}  

\newcommand{\review}[1]{\textcolor[rgb]{0, 0, 0}{#1}}
\newcommand{\tib}{T$^2$IBUA }
\newcommand{\tibb}{T$^2$IBUA}
\newcommand{\myparagraph}[1]{\vspace{2pt}\noindent{\bf #1}}
\newcommand\fontsizenine{\fontsize{9pt}{12pt}\selectfont}
\newcommand{\repeatthanks}{\textsuperscript{*}}

\usepackage{subcaption}

\begin{document}
\pagestyle{headings}
\mainmatter
\def\ECCVSubNumber{10}  

\title{Leveraging Self-Supervised Training for Unintentional Action Recognition} 

\titlerunning{Leveraging Self-Supervised Training for UAR}
%
\author{Enea Duka\thanks{Equal contributions}\and
Anna Kukleva\repeatthanks \and
Bernt Schiele}
\authorrunning{E. Duka et al.}
%
\institute{MPI for Informatics, Saarbrücken, Germany\\
\email{\{enea.duka,akukleva,schiele\}@mpi-inf.mpg.de}\\
\url{https://www.mpi-inf.mpg.de}}
\maketitle

\begin{abstract}
Unintentional actions are rare occurrences that are difficult to define precisely and that are highly dependent on the temporal context of the action. In this work, we explore such actions and seek to identify the points in videos where the actions transition from intentional to unintentional.
We propose a multi-stage framework that exploits inherent biases such as motion speed, 
motion direction, and order 
to recognize unintentional actions.
\review{To enhance
representations via self-supervised training for the task of unintentional action recognition we propose temporal transformations, called 
\textbf{T}emporal \textbf{T}ransformations of \textbf{I}nherent \textbf{B}iases of \textbf{U}nintentional \textbf{A}ctions (\tibb). } 
The multi-stage approach models the temporal information on both the level of individual frames and full clips.  
These \review{enhanced} representations show strong performance for 
unintentional action recognition tasks. 
We  provide an extensive  ablation  study of our framework and report results that significantly improve over the state-of-the-art.
\end{abstract}

\input{1_intro}
\input{2_rel_work}
\input{3_approach}
\input{4_exp_results}
\input{5_conclusions}


\clearpage
%
%
\bibliographystyle{splncs04}
\bibliography{egbib}
\end{document}


\pagestyle{headings}
\mainmatter
\def\ECCVSubNumber{10}  

\title{Leveraging Self-Supervised Training for Unintentional Action Recognition Supplement} 

\titlerunning{Leveraging Self-Supervised Training for UAR Suplement}
%
\author{Enea Duka\thanks{Equal contributions}\and
Anna Kukleva\repeatthanks \and
Bernt Schiele}
\authorrunning{E. Duka et al.}
%
\institute{MPI for Informatics, Saarbrücken, Germany\\
\email{\{enea.duka,akukleva,schiele\}@mpi-inf.mpg.de}\\
\url{https://www.mpi-inf.mpg.de}}
\maketitle

In Sec.~\ref{sec:abl_supmat} of the supplement, we provide further ablation experiments on different components of our framework. Then, in Sec.~\ref{sec:ext_imp_details} we provide additional implementation details for the spatial feature extraction and the self-supervised representation learning training. Then, in Sec.~\ref{sec:limitations} we discuss limitations of our framework. Finally, Sec.~\ref{sec:viz_trn_point} contains qualitative results of our framework.

\section{Additional ablation on framework}
\label{sec:abl_supmat}
In this section, we discuss possible modifications of our framework. Firstly, we explore the influence of different pre-trained backbone models and different methods for initialization.  Further, we evaluate the importance of each transformation to the overall performance. Then, we explore the importance of the first stage of our framework. Further, we assess different transition matrices $T$ for CRF. Moreover, we discuss the influence of the attention window size and the depth of the temporal encoders. Finally, we provide additional evaluation for the anticipation task that extends our study.


\subsection{Backbone model and representation initialization method}

In this section, we explore additional pre-trained frozen backbone models as our spatial encoder $\mathfrak{S}$. We explore ResNet18 (R18), ResNet50 (R50) and ViT architectures as a backbone model for $\mathfrak{S}$. As initialization task for all the backbones we consider fully-supervised (FS) image classification on ImageNet~\cite{deng2009imagenet} and ImageNet 21K~\cite{ridniki2021_21k}. 
We additionally include R3D with ResNet18 backbone (R(18)3D) from the main paper that we use for a fair comparison with the previous methods that we train from a random initialization. Note that the spatial pre-trained models we keep frozen while R(18)3D we train during all the stages.  In Tab.~\ref{tab:backbone_finetune} we show results for both F2C and F2C2V stages of our framework. 
By using frozen pre-trained convolution backbones R18 and R50 we can notice similar performance with F2C2V learning with the difference in 0.5 points while with F2C learning the gap is 0.9 points. 
That indicates that the global F2C2V feature enhancement closes the gap between different frozen spatial features. 
Further, we can observe that transformer based architecture (ViT) improves the performance by 2 points for the second stage. 
In Tab.~\ref{tab:backbone_finetune} we show that the difference between different backbones is significant for F2C learning stage that encodes local clip information, the gap between R18 (IN 1K) and ViT (IN 21K) is 5.2 points. Whereas for F2C2V learning stage the gap is only 2.5 points that shows the importance of the two stage feature enhancement. 

\input{tables/uar_ablation_backbone_finetune}


\subsection{Influence of temporal transformation}
In this section, we evaluate the influence of each transformation on the overall performance. For this experiment, we selectively remove each transformation from Stage 1 and Stage 2 of our framework. In Tab.~\ref{tab:trn_ablation} we show the overall performance and the loss for each transformation exclusion. We consider the most influential transformation to be that corresponding to the greater loss in performance. We notice that the most influential transformation is Random Point Speedup in both F2C and F2C2V stages. In addition, we notice that all the transformations contribute to the overall performance of the framework.
\input{tables/uar_trn_ablation}

\subsection{Importance of Frame2Clip learning [Stage 1]}
In this section, we assess the influence of the first learning stage of our framework on the overall performance. First, we skip the first stage where we pretrain the F2C temporal encoder, and instead, we directly pretrain the overall temporal encoder F2C2V on self-supervised representation learning on the clip level. Specifically, we jointly pretrain both $\mathfrak{T}_{frame}$ and $\mathfrak{T}_{clip}$ during the second stage of our learning protocol and show the performance in Tab.~\ref{table:uar_ablation_stage1} in the first row.
Next, we replace $\mathfrak{T}_{frame}$ with temporal global average pooling (GAP) and report the performance in the second row. Finally, in the last row, we observe that separate stages of learning for $\mathfrak{T}_{frame}$ and $\mathfrak{T}_{clip}$ lead to the best overall performance. 

\subsection{Importance of CRF layer}
In this section, we evaluate the influence of CRF on the performance. In Tab.~\ref{table:uar_ablation_crf_influence} the last row shows the performance of the model for F2C and F2C2V temporal encoders with CRF layer on the top. We compare the performance to the second row, with \tib and without CRF, and we can observe that for the F2C there is no improvement, while for F2C2V we obtain $2.6$ points increase compared to the model without CRF. We assume that inter clip temporal information from the whole video helps to encode global information better and thus to recognize UA with better precision. Additionally, we depict the learned transition matrix of the CRF layer in Fig.~\ref{fig:trn_matrix}. We can notice the preference of the strict ordering from intentional to transitional and to unintentional clips. 

\begin{table}[h!]
\centering
\begin{minipage}{.4\textwidth}
  \centering
   \caption{Influence of Stage 1 on UA classification task. GAP denotes global average pooling operator instead of the $\mathfrak{T}_{frame}$ encoder. }
    \begin{tabular}{c|c}
    \toprule
    Stage 1 & Accuracy \\
    \midrule
    $-$& 74.4\\
    GAP & 75.0\\
    $\mathfrak{T}_{frame}$ & \textbf{76.9}\\
    \bottomrule
    \end{tabular}
    \label{table:uar_ablation_stage1}
\end{minipage}%
\hfill
\begin{minipage}{.58\textwidth}
  \centering
  \caption{\fontsizenine{Ablation results: Influence of \tib and CRF layer on classification performance for two representations, pretrained ViT representation and R3D.}}
    \label{table:uar_ablation_crf_influence}
    \begin{tabular}{c|c|cc|cc}
    \toprule
    \multirow{2}{*}{\tib} & \multirow{2}{*}{CRF} & \multicolumn{2}{c}{ViT}  & \multicolumn{2}{c}{R3D} \\
    & & F2C & F2C2V & F2C & F2C2V \\ 
    \midrule
    $-$ & $-$ &  60.9 & 69.6 & 59.3 & 65.6\\
    \ding{52}&$-$ &  65.5 & 74.3 & 63.8 & 70.4 \\
    \ding{52}&\ding{52} & 65.0 & \textbf{76.9} & 65.3 & \textbf{74.0}\\
    \bottomrule
    \end{tabular}

\end{minipage}
\end{table}

\subsection{CRF transition matrix calculation}
\begin{figure*}[ht!] 
    \centering
    \subfloat[Binary]{%
        \includegraphics[width=0.3\textwidth]{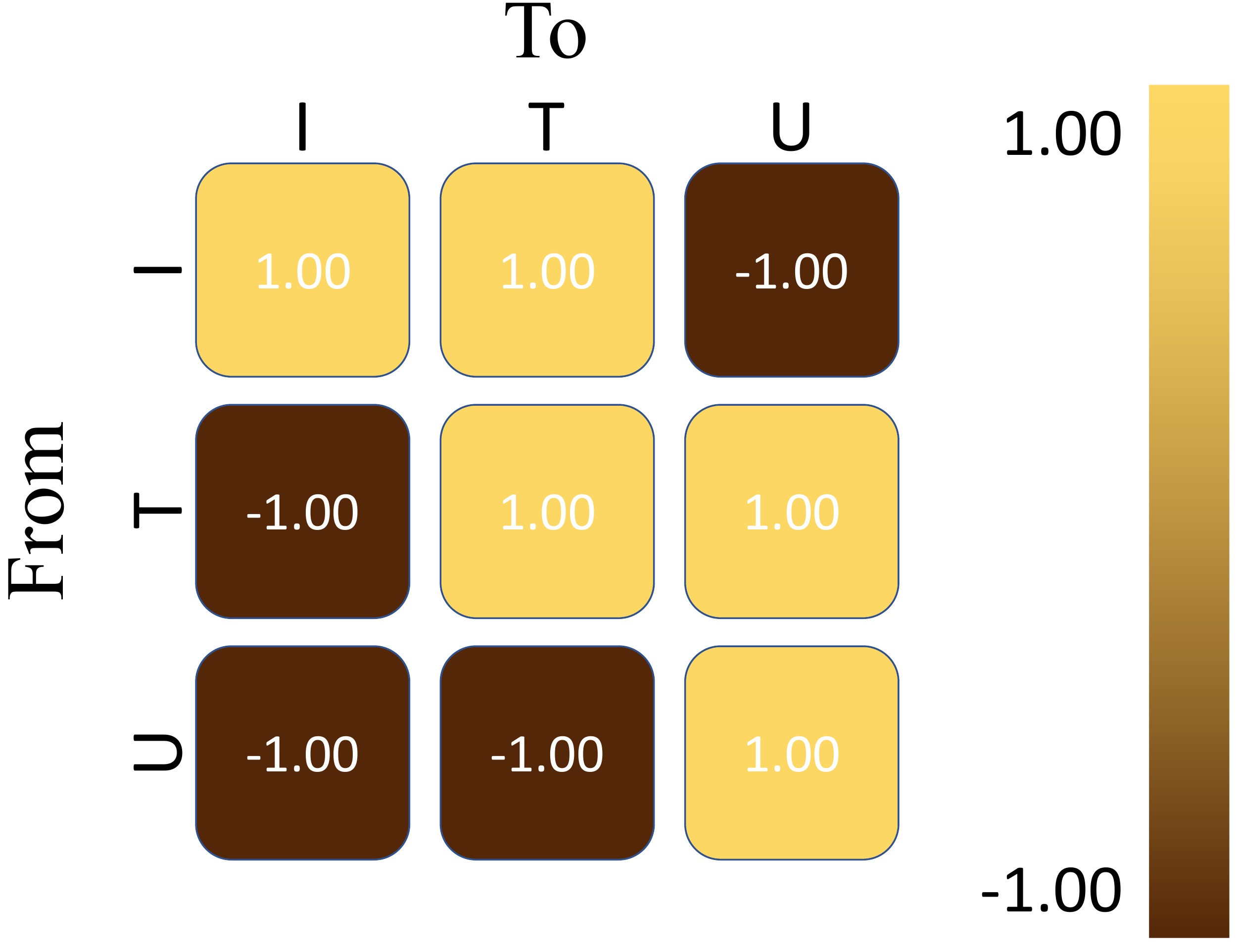}%
        \label{fig:trn_matrix_binary}%
        }%
    \hfill%
    \subfloat[Prior statistic from training split]{%
        \includegraphics[width=0.3\textwidth]{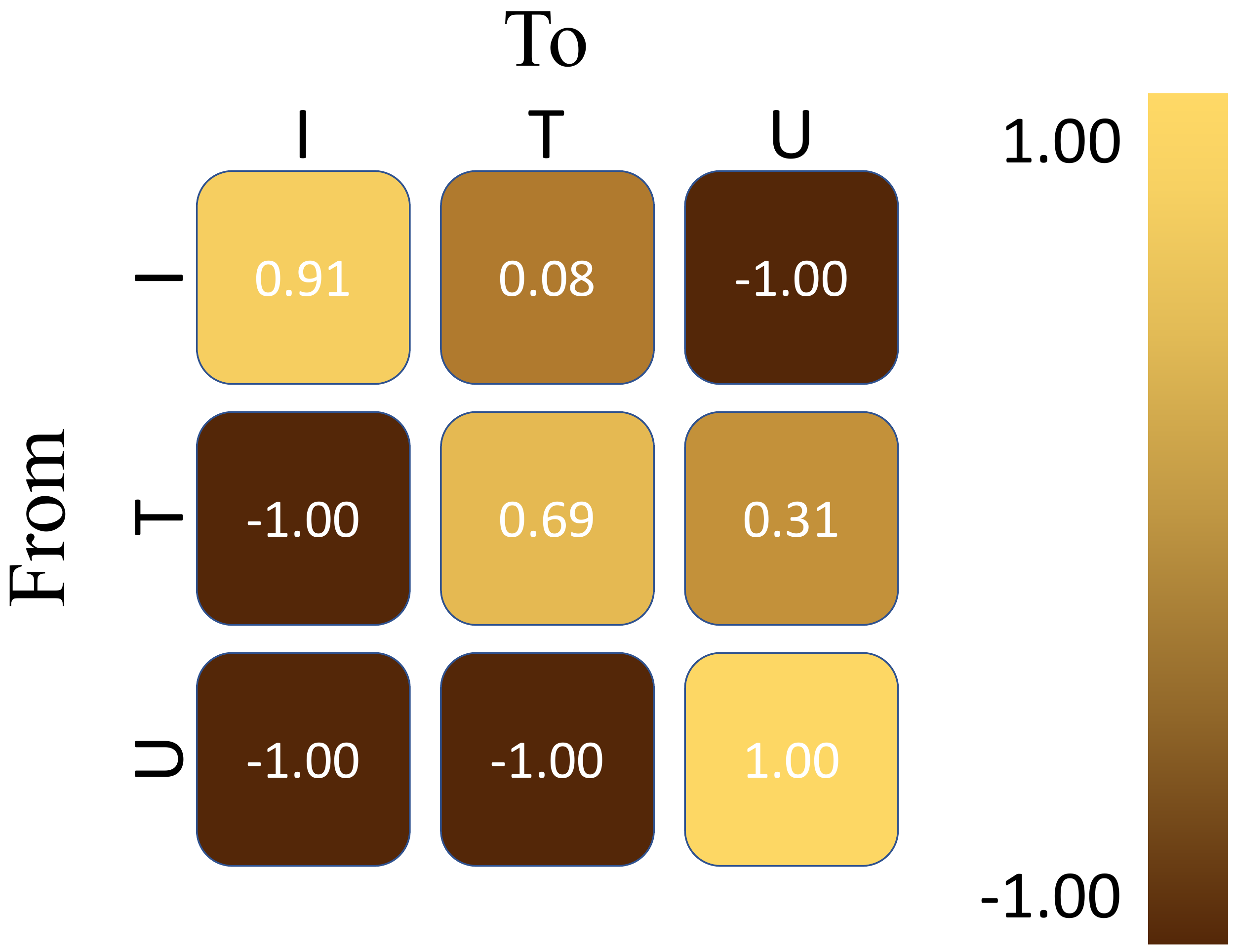}%
        \label{fig:trn_matrix_count}%
        }%
    \hfill%
    \subfloat[Trainable weights]{%
        \includegraphics[width=0.32\textwidth]{figures/trn_matrix.pdf}%
        \label{fig:trn_matrix_opt}%
        }%
    \caption{Different approaches for calculating the transition score matrix. I: Intentional, T: Transitional, U: Unintentional actions.}
\end{figure*}

In this section, we evaluate different methods to calculate the weights of the CRF transition matrix $T$ and show the performance  in Tab.~\ref{table:uar_ablation_trn_matrix}. First, we define a binary transition matrix where $T_{i, j}=1$ denotes that the transition from class $i$ to class $j$ is possible and otherwise for $T_{i, j}=-1$. See Fig.~\ref{fig:trn_matrix_binary} for the visualization of the binary transition matrix. Next, we calculate the weights for the transition matrix based on a statistic from the training set, see Fig.~\ref{fig:trn_matrix_count}. For each element $T_{i, j}$ we count the transitions from class $i$ to $j$ in the training split of the dataset and then normalize each row to form a probability distribution. We replace $0$ with $-1$ to further discourage during inference the transitions that do not occur in the training split. Finally, we optimize the weights during the training process, as we discuss in the main paper. The resulting matrix is shown in Fig.~\ref{fig:trn_matrix_opt}. We notice that the transition matrix $T$ with the trainable weights leads to the best performance for our framework.

\begin{table}
\vspace{-4mm}
\centering
\begin{minipage}{.48\textwidth}
  \centering
  \includegraphics[width=.8\linewidth]{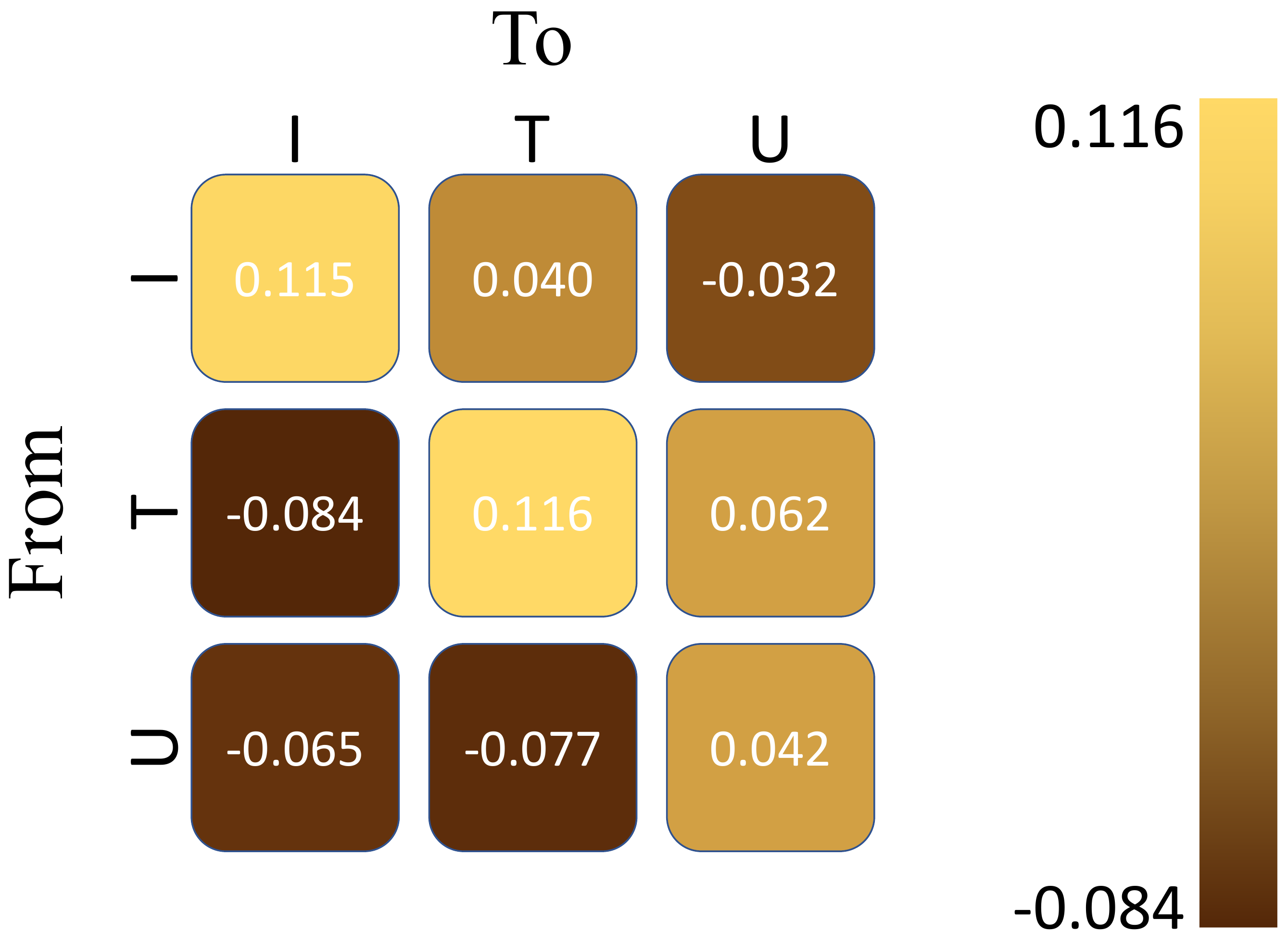}
  \captionof{figure}{\fontsizenine{Learned transition score matrix for the CRF layer.   I: Intentional, T: Transitional, U: Unintentional actions.}}
  \label{fig:trn_matrix}
    \end{minipage}%
\hfill
\begin{minipage}{.48\textwidth}
  \vspace{-2mm}
  \centering
   \caption{Influence of transition score matrix calculation method on UA classification task.}
    \label{table:uar_ablation_trn_matrix}
    \begin{tabular}{c|c}
    \toprule
    Calc. Method & Accuracy \\
    \midrule
    Binary & 74.8\\
    Training split prior & 74.9\\
    Trainable weights & \textbf{76.9}\\
    \bottomrule
    \end{tabular}
\end{minipage}
\vspace{-4mm}
\end{table}

\subsection{Attention mechanism window size}
In this section, we evaluate the influence of the attention window size on performance. In Tab.~\ref{table:uar_ablation_win_size} we compare the influence on both F2C and F2C2V stages. The input for the first stage encoder F2C is a clip of a fixed length, and we can observe a significant influence of the window size on the performance. Note that the performance is evaluated for the first stage of our temporal encoder. We obtain the best performance with the full attention mechanism, where the window size is equal to the length of the sequence. Whereas during the second stage, the input to the F2C2V temporal encoder is a sequence of clips of arbitrary length, since all the videos are of different lengths. For the two-stage temporal encoder F2C2V, we notice smaller variations in performance. We find that in both cases the optimal window size is $32$. 
\input{tables/uar_ablation_win_size}
\vspace{-4mm}
\vspace{-4mm}

\subsection{Anticipating further into the future} 
In this section, we evaluate our framework anticipating unintentional actions. We anticipate with different time points $\tau_A$ in the future. In Tab.~\ref{table:uar_ablation_anticipation} we notice an overall increasing trend, but this does not necessarily mean that our framework anticipates actions better further into the future. The trend we notice is because the number of unintentional action class samples increases the further we anticipate into the future. We can conclude that our framework anticipates actions further into the future with a similar accuracy it classifies current actions.
\input{tables/uar_ablation_anticipation}

\subsection{Depth of the temporal encoder} We explore how the depth of the temporal encoder influences the performance. In Tab.~\ref{table:uar_ablation_lnf_layers} we notice that a depth of 3 layers works the best. Less or more layers lead to a decrease in performance. This might be due to the capacity of the framework being too low and too high respectively.

\input{tables/uar_ablation_lnf_layers}

\section{Extended implementation details}
\label{sec:ext_imp_details}
In this section, we discuss additional details of the different stages of our framework. 

\myparagraph{Spatial feature extraction} We extract spatial features for each frame of the videos separately. Each frame is scaled so that the smallest dimension is $224$ pixels. After scaling, we perform center crop to have $224 \times 224$ pixel frames.

\myparagraph{Self-supervised training} During the first and second stages, we sample one of the following ratios $\{ \nicefrac{1}{2}, \nicefrac{1}{4}, \nicefrac{1}{8} \}$ with which we subsample the video frames when applying the \emph{Speed-up} transformation. We first filter these ratios according to the video length, so that the transformed video is at least $48$ frames long for the first stage and $3$ clips long for the second stage. Then, from the filtered ratios, we randomly sample one of them. During the first stage, we apply temporal augmentation as follows. After each transformation is applied to the input video, and we have $6$ transformed versions of it, we crop the video versions randomly so that the final length is between $75\%$ and $100\%$ of the original length of each version. This is used as a temporal augmentation technique during this stage. 

\section{Limitations}
\label{sec:limitations}
Due to the diversity of unintentional actions, our approach does not cover every aspect of unintentionality resulting in powerful but suboptimal representations. For unintentional action localisation, we rely on the classification confidence of our framework. While our classification accuracy is high, the localisation accuracy does not follow. Due to sampling of the overlapping clips, we predict the transitional point less reliably. 
Finally, the CRF layer significantly advances the performance of our model based on the fact that each video contains only one transitional point, we assume that more frequent transition points would lower performance notably.
We notice these limitations of our current work, and we plan to address them in the future.

\section{Qualitative results}
\label{sec:viz_trn_point}
\begin{figure*}[h]
\centering
\includegraphics[width=0.9\linewidth]{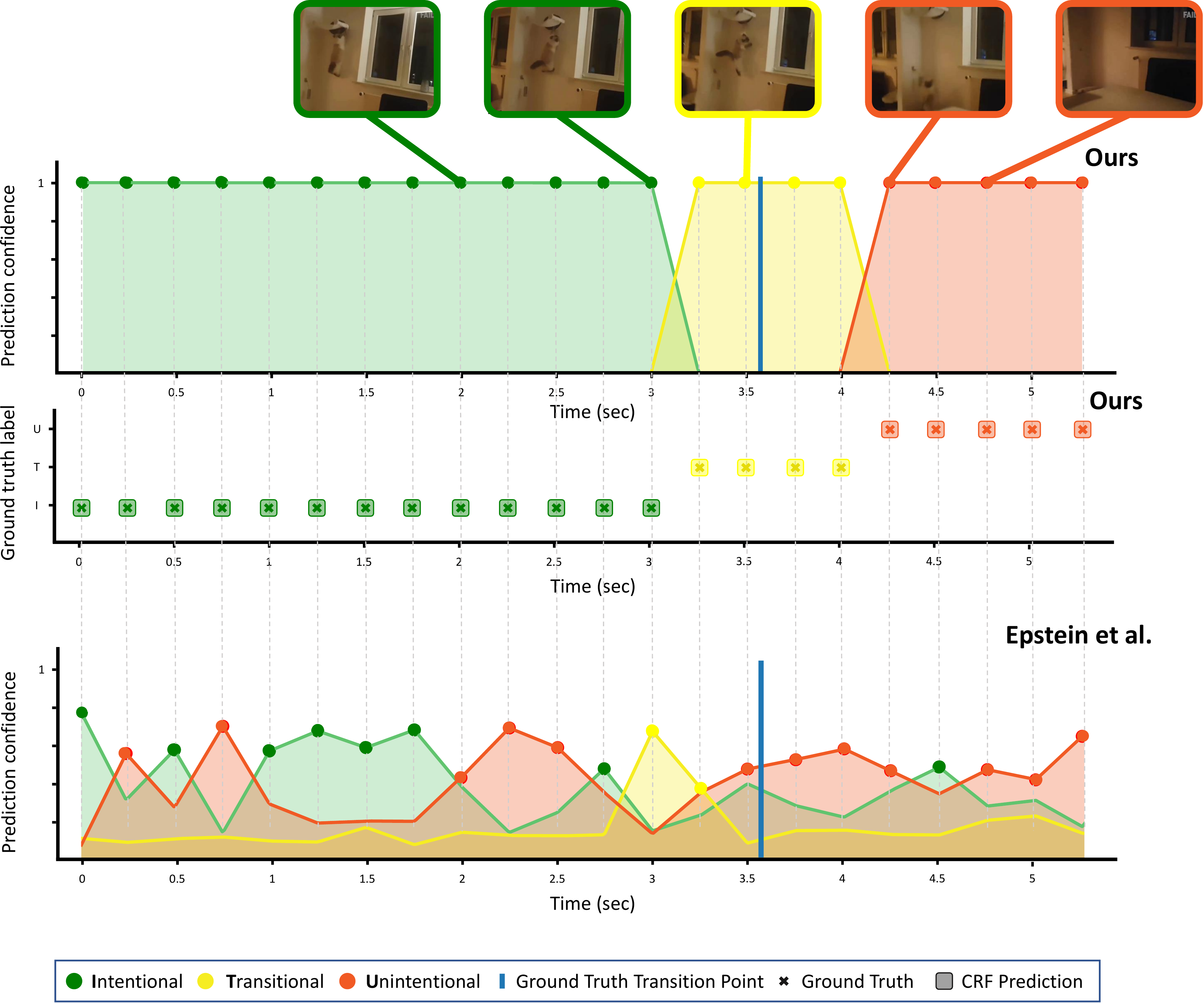}
\caption{A cat falling while trying to hold on to an object on the wall. We notice that when we do not use CRF, all the predictions are correct and with high confidence. After adding CRF, the predictions remain all correct.}
\label{fig:qual_best_one}
\vspace{-3mm}
\end{figure*}

\begin{figure*}[ht!]
\centering
\includegraphics[width=0.9\linewidth]{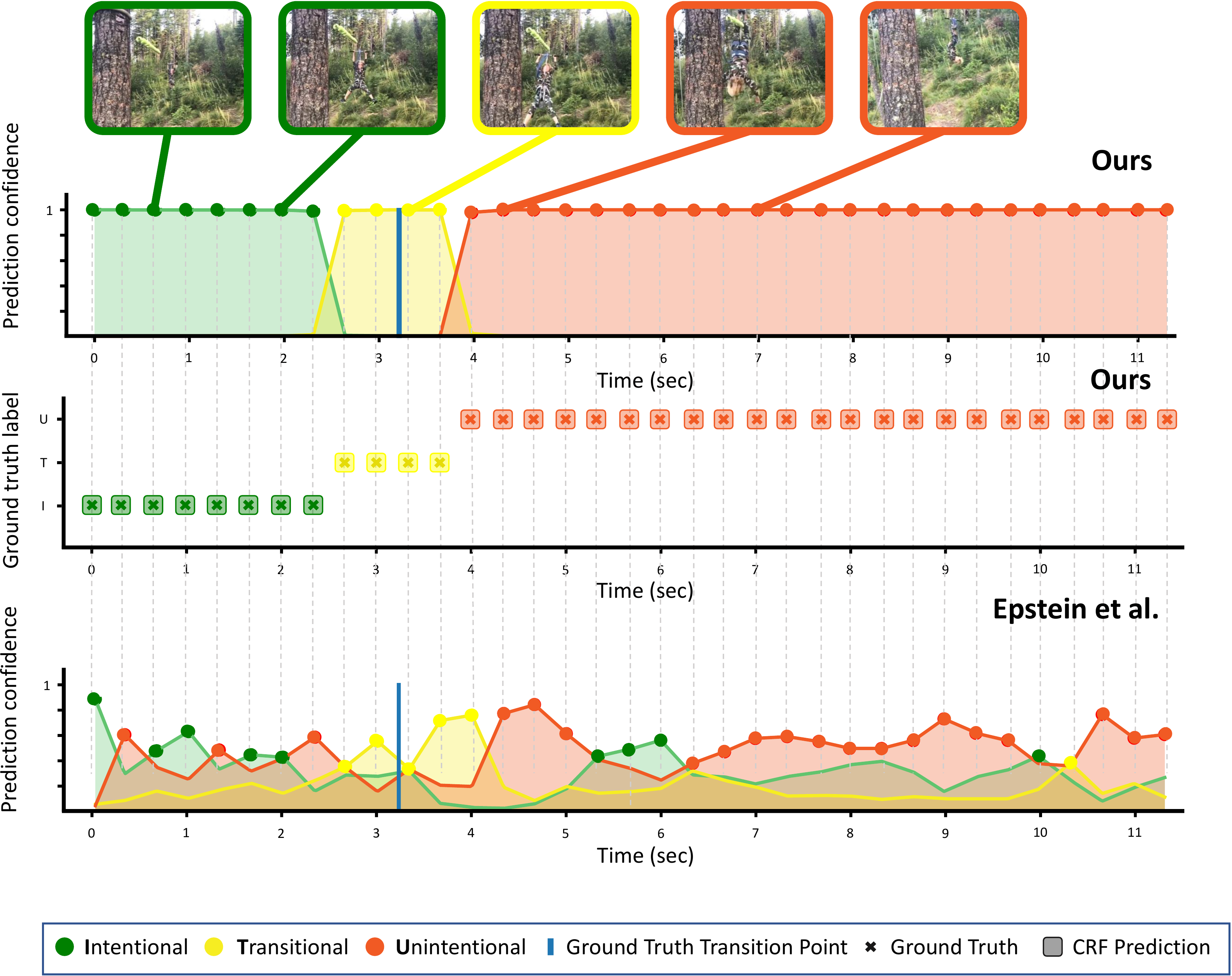}
\caption{A person abruptly ending the zipline ride and turning upside-down at the end of it. We notice that when we do not use CRF, all the predictions are correct and with high confidence. After adding CRF, the predictions remain all correct.}
\label{fig:qual_best_two}
\vspace{-3mm}
\end{figure*}

\begin{figure*}[ht!]
\centering
\includegraphics[width=0.9\linewidth]{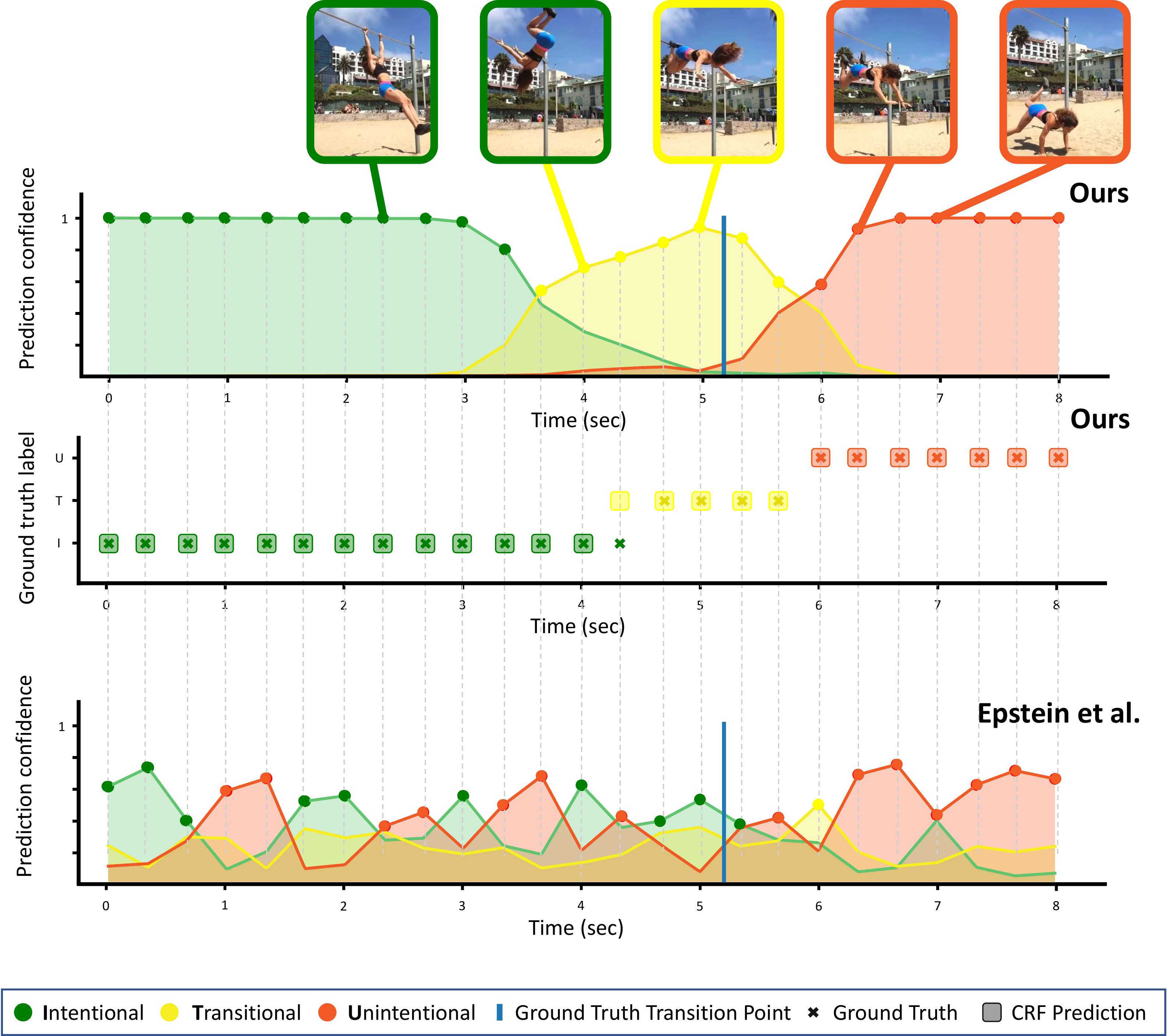}
\caption{A person trying to perform tricks on a pull-up bar and falling down in the sand. We notice that when we do not use CRF, most of the predictions are correct and not all of them have a high confidence. Adding CRF in this case leads to more predictions being correct.}
\label{fig:qual_good_one}
\vspace{-3mm}
\end{figure*}

\begin{figure*}[ht!]
\centering
\includegraphics[width=0.9\linewidth]{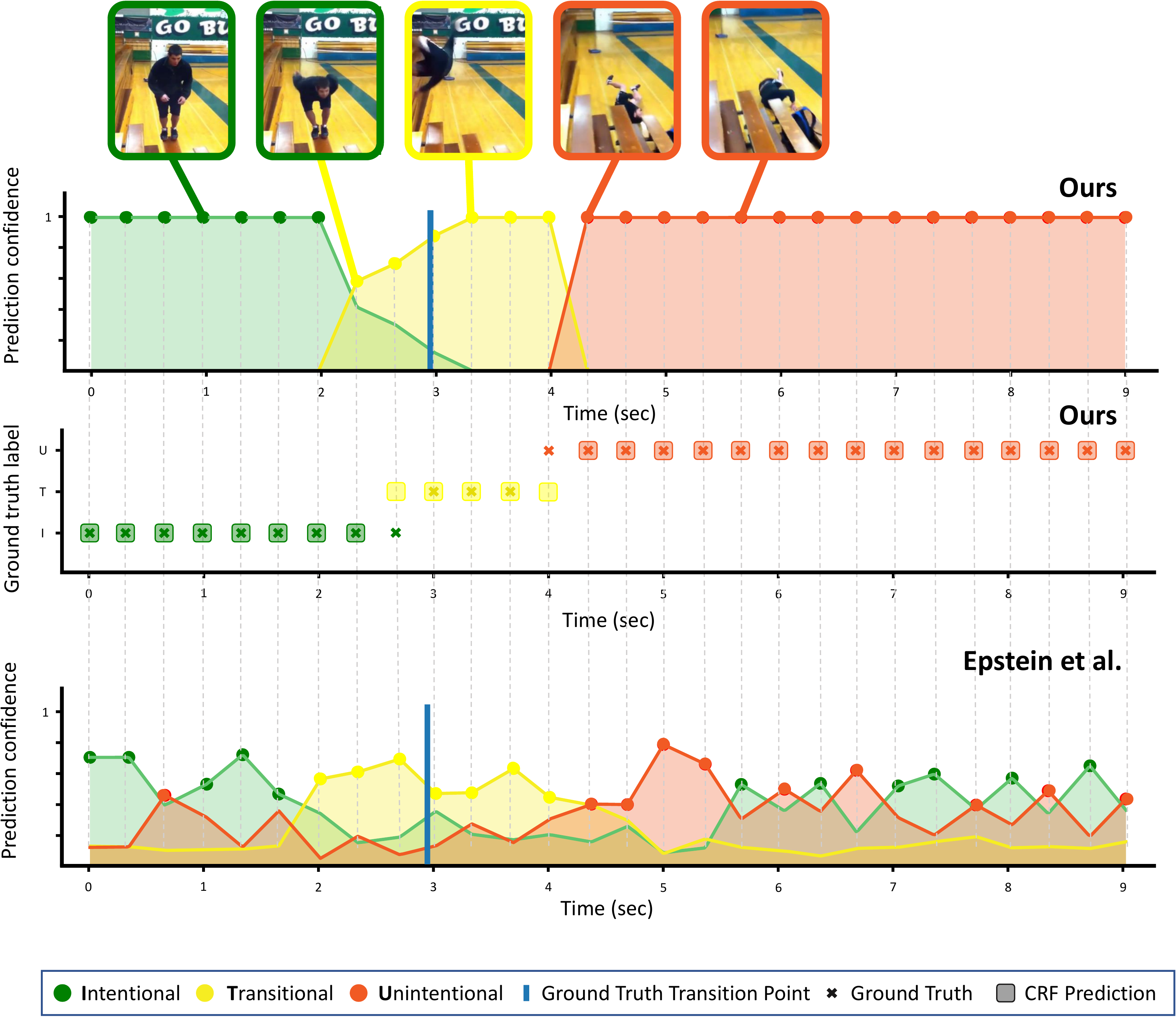}
\caption{A person trying to perform a backflip from the top of a bench and falling on the floor. We notice that when we do not use CRF, most of the predictions are correct and not all of them have a high confidence. Adding CRF in this case leads to marginally better performance.}
\label{fig:qual_good_two}
\vspace{-3mm}
\end{figure*}

\begin{figure*}[ht!]
\centering
\includegraphics[width=0.9\linewidth]{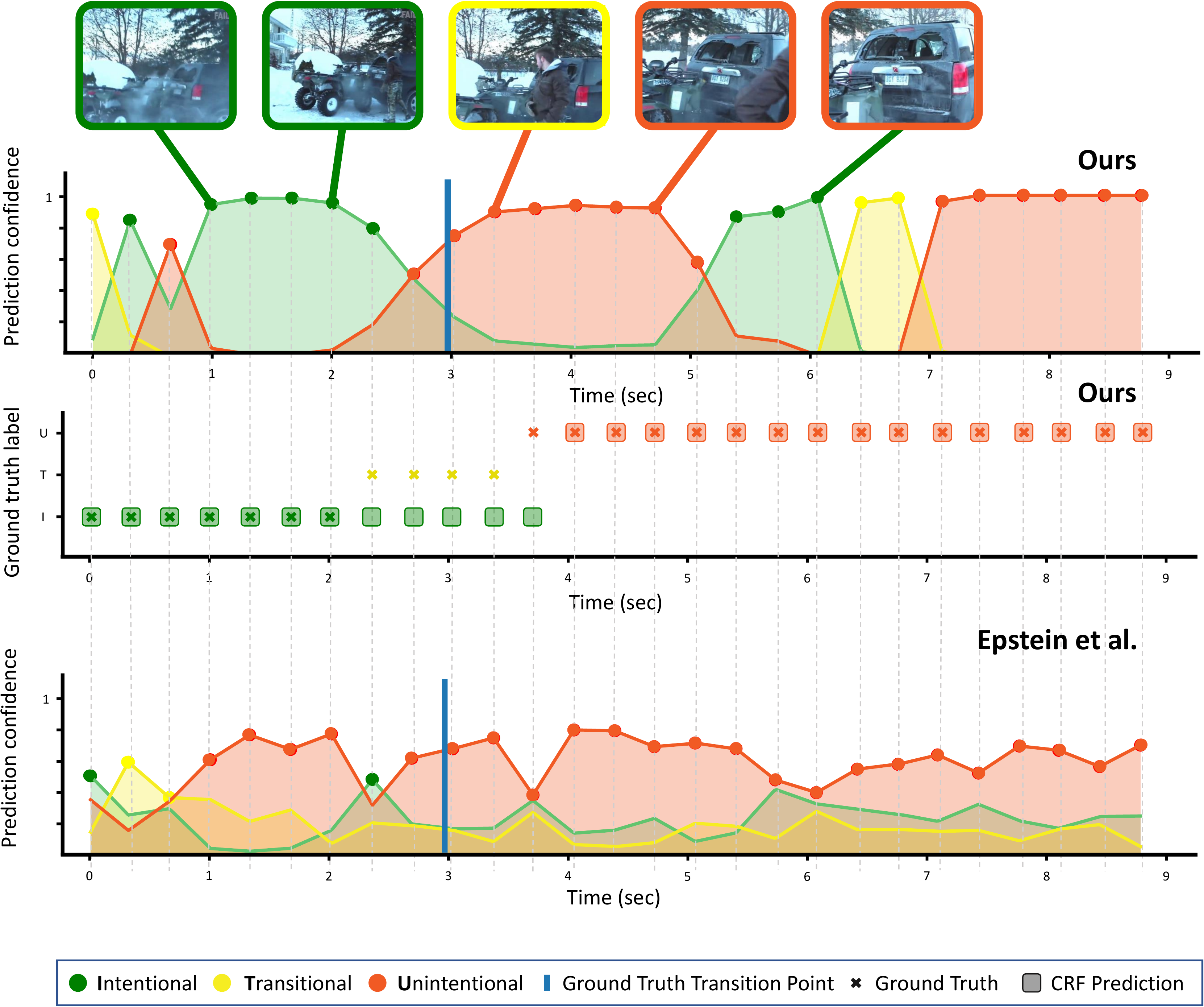}
\caption{An ATV has hit the back of a car and the camera films the aftermath. We notice that when we do not use CRF, there are few predictions that are correct, and they are noisy overall. In addition, it is hard to correctly locate the transition point from intentional to unintentional action. Adding CRF makes the predictions overall smoother and less noisy. However, it is still hard to locate the transition point.}
\label{fig:qual_bad_one}
\vspace{-3mm}
\end{figure*}

\begin{figure*}[ht!]
\centering
\includegraphics[width=0.9\linewidth]{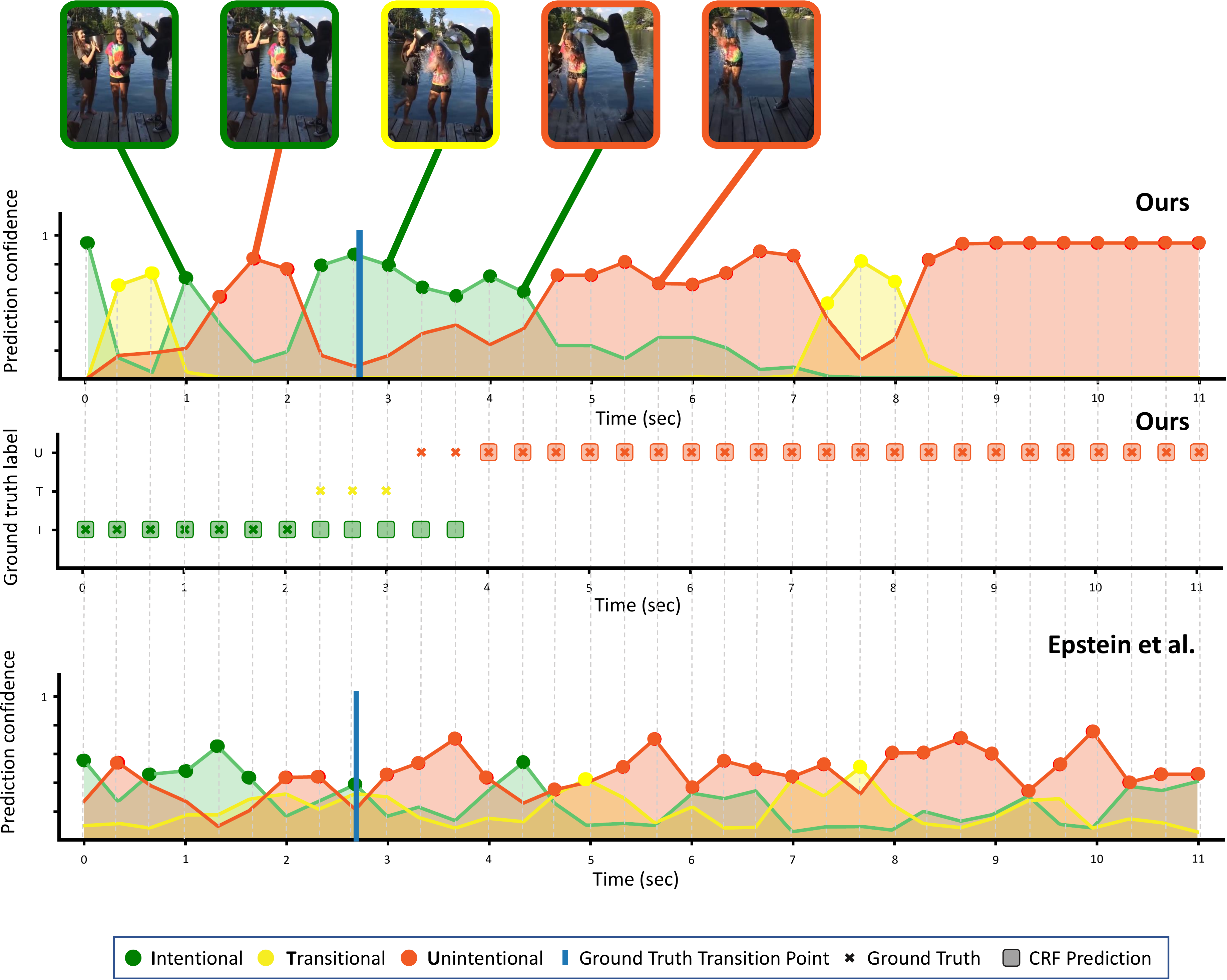}
\caption{The person on the left falling in the lake while trying to throw a bucket of water on the person in the middle. We notice that when we do not use CRF, there are few predictions that are correct, and they are noisy overall. In addition, it is hard to correctly locate the transition point from intentional to unintentional action. Adding CRF makes the predictions overall smoother and less noisy. However, it is still hard to locate the transition point.}
\label{fig:qual_bad_two}
\vspace{-3mm}
\end{figure*}

In this section, we show qualitative results for our framework. Fig.~\ref{fig:qual_best_one},~\ref{fig:qual_best_two},~\ref{fig:qual_good_one},~\ref{fig:qual_good_two},~\ref{fig:qual_bad_one},~\ref{fig:qual_bad_two} display these results. In each case, the first plot in the figure shows the confidence of our framework before using CRF. The $x$ axis represents the time in seconds, while the $y$ axis represents the confidence for that prediction. We mark the ground truth transition point $t_{gt}$ of the video with the vertical line at $x=t_{gt}$. The second plot in each figure shows the predictions of the framework when we include CRF. The $x$ axis shows the time in seconds, while the $y$ axis shows the discrete clip label. The third plot in each figure, shows the shows the confidence of model in \cite{epstein2019unint_actions} and has the same layout as the first plot. At the top of the figure, we show frames for clips close to the transition point in the video. The color of the frame border indicates the ground truth label for it. We take the frame in the middle of the clip each prediction is related to.  

In Fig.~\ref{fig:qual_best_one},~\ref{fig:qual_best_two}, we show samples for which our network performs the best. First, we notice in the continuous plot that the prediction is correct for each clip. In addition, we can see that the confidence is high for all clips and there is a clean transition point and ordering between the different types of clips. These results translate to the discrete scatter plot, where we notice that all the predictions are correct. In contrast to our results, the results from~\cite{epstein2019unint_actions} are clearly more noisy and less correct.

In Fig.~\ref{fig:qual_good_one},~\ref{fig:qual_good_two} we show samples for which the predictions are less accurate. We observe that the order of the clip types is preserved. However, the exact transition from one clip type to the next is less certain, specifically, the confidence of the predictions on the boarders between the different types is lower than the maximal possible score as in the previous case. The same is not true for results from~\cite{epstein2019unint_actions} where not only the prediction confidence is suboptimal but also the clip order is not preserved.
In the scatter plots of both figures, we notice the improvement on the performance due to the CRF layer.

Finally, in Fig.~\ref{fig:qual_bad_one},~\ref{fig:qual_bad_two} we show samples on which our framework performs the worst. In this case, we observe that there are noisy predictions which violate the order of the clip types as well as the clean transitions between the clip types. We notice similar results for these samples when using the model from~\cite{epstein2019unint_actions}. CRF layer improves the quality of the predictions in these cases the most. It makes the prediction smoother and reduces the noise, but the transition point localisation remains poor.

\clearpage
%
%
\bibliographystyle{splncs04}
\bibliography{egbib}

%% file: 1_intro.tex
\section{Introduction}
\label{sec:intro}

Video action understanding has witnessed great progress over the past several years in the fields of action detection, recognition, segmentation, caption generation, tracking and many others~\cite{dai2021dialated_pyramid_network},~\cite{yang2020pyramid_network},~\cite{kukleva2019unsup_act_seg},~\cite{ging2020cap_gen},~\cite{wang2021trn_tracking}. However, these methods implicitly rely on the continuity of the underlying intentional action, 
such as one starts and then continues the same or related activity for some time.
In this paper, we study unintentional actions, for example, when one falls down by accident during jogging. This is a challenging task due to the difficulty in defining precisely the intentionality of an action while largely depending on the temporal context of the action. Moreover, annotating such videos is both costly and difficult with human accuracy being 88\% in localising the transition between intentional and unintentional actions~\cite{epstein2019unint_actions}. The potential of this research covers assistance robotics, health care devices, or public video surveillance, where detecting such actions can be critical. In this work, we propose to look at unintentional actions from the perspective of 
the discontinuity of the intentionality 
and exploit inherent biases in this type of videos.

We design our framework to explore videos
where a transition from intentional to unintentional action occurs.
We are the first to explicitly leverage the inherent biases of unintentional videos in the form of motion cues
and high-level action ordering within the same video, specifically the transition from unintentional to intentional and vice-versa.


\begin{figure}[t!]
\centering
\includegraphics[width=0.5\linewidth]{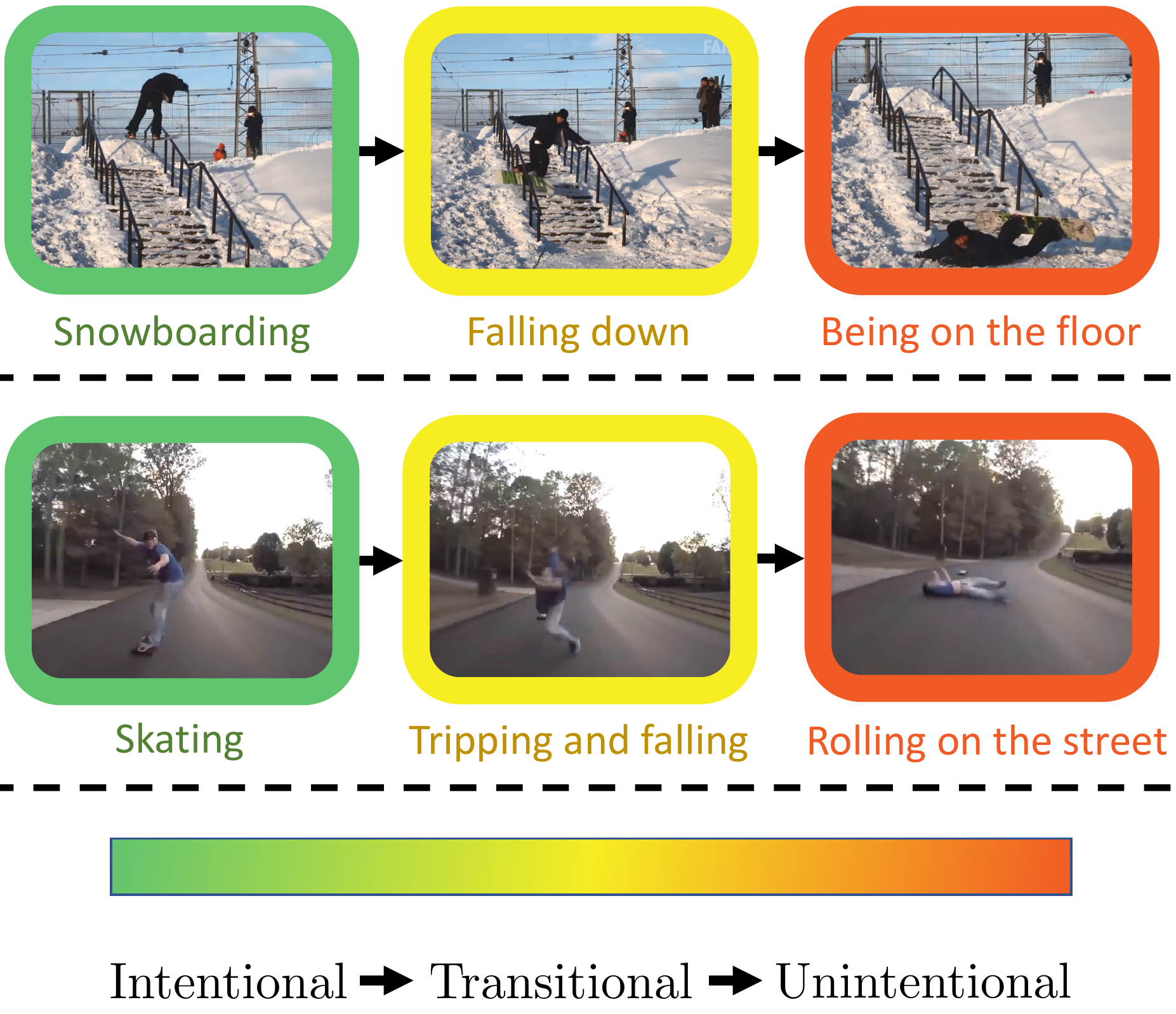}
\caption{\fontsizenine{Transition from intentional to unintentional action. We can notice an abrupt change of motion in the first example, when the snowboarder starts to fall down the stairs, and a sudden change in speed in the second example when the skater starts rolling on the ground. In both cases, there is a strict order from intentional to unintentional action.}}
\label{fig:transition_frames_fig}
\end{figure}

To this end, we propose a three-stage framework where, starting from a pre-trained representation, the first two stages \review{leverage self-supervised training} to address unintentional biases on frame and video clip levels \review{and further enhance these representations.}
We regularly observe abrupt motion changes in videos of unintentional actions when an unwitting event happens. 
Therefore, we formulate \textbf{T}emporal \textbf{T}ransformations to exploit \textbf{I}nherent \textbf{B}iases in videos that resemble \textbf{U}nintentional \textbf{A}ction motion changes (\tibb).
Using \tib we learn intra and inter-clip information in  a multi-stage manner. In the first stage, we capture local information from the neighbouring frames by predicting the discrete label of the applied \tib on the frame level. 
Then, in the second stage, we train the model to integrate information from the entire video by predicting \tib on the clip level. See Fig.~\ref{fig:frame_trc_2_clip_trn} for different \tib levels. Inspired by the growing popularity of transformer models in vision~\cite{fan2021mvit},~\cite{sun2019video_bert},~\cite{girdhar_act_trn_net},~\cite{neimark2021vtn}, we utilize a multi-level transformer architecture that serves as a temporal encoder in our framework to model long-term relations. 
In the last stage of supervised downstream learning, we particularly benefit from the high-level ordering by deploying conditional random fields (CRF) to explicitly enforce smooth transitions within the same video from intentional to unintentional actions. By modelling these explicit global temporal relations, we enhance our results on various downstream tasks such as unintentional action classification, localisation and anticipation.

Our work makes the following contributions. Firstly, we propose a framework that includes three learning stages for unintentional action recognition: 
\review{by leveraging inherent biases in the first and in the second stages with self-supervised feature enhancement,
the third stage employs supervised learning for unintentional action recognition on the top of these amplified representations. }
In addition, we present a multi-stage temporal encoder as a transformer-based architecture in combination with conditional random fields to capture local and global temporal relations between intentional and unintentional actions. Finally, we show state-of-the-art results for classification, detection, and anticipation tasks and perform various ablation studies that evaluate the effectiveness of each of the components of the proposed framework.


%% file: 2_rel_work.tex
\section{Related Work}
\label{sec:rel_work}
 \myparagraph{Unintentional Action Recognition.}
 The topic of unintentional action recognition is recently introduced by Epstein et al.~\cite{epstein2019unint_actions}. The authors collect the dataset Oops! and propose a framework to study unintentional action recognition based on three directions such as classification, localization, and anticipation. The framework consists of two consecutive parts to learn the representations in a self-supervised way and then adapt the model to unintentional actions.  In contrast to~\cite{epstein2019unint_actions} we  \review{leverage self-supervised training to tailor pre-trained representations to} unintentional actions and model temporal information in a multi-stage fashion through the multiple learning stages. 
Han et al.~\cite{han2020mem_dpc} predict the future steps of the video based on the feature space on the given history frames.
Epstein et al.~\cite{epstein2020gloal_from_failure} utilize a 3D CNN in combination with Transformer blocks to enhance representation learning and recognize discontinuities in the videos. As opposed to~\cite{epstein2020gloal_from_failure} we follow a two-step approach with a \review{ self-supervised enhancement of the pre-trained features} followed by a fully-supervised fine-tuning on unintentional action recognition tasks as in~\cite{epstein2019unint_actions},~\cite{han2019dpc}. 

 \myparagraph{Video Representation Learning.} 
 Nowadays, video representation learning is an active research topic. Various types of architectures are developed, such as two-stream networks 
 \cite{carreira2017i3d},~\cite{feichtenhofer2016two-stream_fusion},~\cite{chen2019improved_two-stream}, 3D convolution networks 
 ~\cite{carreira2017i3d},~\cite{tran2014c3d},~\cite{tran2017r(2+1)d}, Transformer based architectures
 ~\cite{liu2021swin_transformer},~\cite{neimark2021vtn},~\cite{fan2021mvit} that rely on supervised training. Many other works explore self-supervised learning to utilize more widely available resources~\cite{jayaraman2016ego_motion},~\cite{agrawal2015learn_from_move},~\cite{kim20183d_puzzle},~\cite{vondrick2016generating_scenes}. 
  The first methods~\cite{misra2016seq_verification_rep_lrn},~\cite{lee2017sorting_sequences},~\cite{fernando2016odd_one_out} focus on frame-level video augmentations such as frame shuffling and odd frame insertion in an existing frame sequence. Then they train their model to distinguish these augmentations~\cite{misra2016seq_verification_rep_lrn}.
  Xu et al. ~\cite{xu2019clip_order_pred} modifies this paradigm by augmenting the videos on the clip level instead of frame level. The authors shuffle the video clips and predict the correct order of the clips, thus they employ spatial-temporal information from videos. Han et al.~\cite{han2019dpc} formulate the problem as future clip prediction. The recent work~\cite{cont_video_rep_lrn},~\cite{dave2021temp_contrastive_learning} apply contrastive learning between different augmentation of videos. Some other directions for the self-supervised learning include colourization of greyscale images as a proxy task~\cite{larrson2017color_proxy},~\cite{vondrick201color_tracking}, or leveraging temporal co-occurrence as a learning signal~\cite{isola2015coocurances},~\cite{wang2015unsupervised_learning_vid_rep}. Wei et al.~\cite{wei2018arrow_of_time} utilize the arrow of time by detecting the direction of video playback. Another popular direction for the video representation learning is to learn temporal or spatial-temporal cycle-consistency as a proxy task~\cite{wang2019correspondence},~\cite{jabri2020spat_temp_correspondence}. While in this work, we introduce a \review{self-supervised feature enhancement} that exploits the inherent biases of unintentional actions. \review{We show that this amplification improves the unintentional action recognition by large margin with respect to previous work and the baselines. } 
\label{subsec:vad_rel_worlk}
\label{subsec:kt_rel_worlk}


%% file: 3_approach.tex
\section{Approach to Exploit the UA Inherent Biases}
\label{sec:approach}

In this and the next sections, we present our framework to study unintentional actions (UA) in videos. First, we provide an overview of our approach in Sec.~\ref{subsec:framework}. In Sec.~\ref{subsec:trans} we detail \tib for self-supervised training, and then in Sec.~\ref{sec:stages} we describe the learning stages for our framework. 

\myparagraph{Notation:} Let $X \in \mathcal{R}^{T \times W \times H \times 3}$ be an RGB video, where $T, W$ and $H$ are the number of frames, width and height respectively. We denote a clip sampled from this video as $x \in \mathcal{R}^{t \times W \times H \times 3}$ where $t$ is the number of frames in the clip and $t \leq T$. As $f$ we further denote individual frames from the video or clip. We denote \tib  
as a function $\mathcal{T}(\cdot)$, the spatial encoder as $\mathfrak{S}(\cdot)$ and the temporal encoder as $\mathfrak{T}(\cdot)$, and a linear classification layer as $MLP_i(\cdot)$, where $i$ is a learning stage indicator. 

\subsection{Framework Overview}
\label{subsec:framework}

Unintentional actions (UAs) are highly diverse and additionally happen rarely in daily life, making it difficult to collect representative and large-scale datasets. To overcome this issue, we propose a three-stage learning framework for unintentional action recognition. Starting from the observation that unintentional actions are often related to changes in motion such as speed or direction, we aim to leverage such inherent biases of unintentional actions in our framework. In particular, simulating these inherent biases with a set of temporal video transformations, the first two stages of our framework \review{greatly enhance pre-trained representation} in a self-supervised fashion.
More specifically, the first stage uses these transformations to learn intra-clip information (frame2clip) whereas the second stage uses the same transformations to address inter-clip information (frame2clip2video). The third stage then refines this representation via fine-tuning on the downstream unintentional action tasks using labelled data. We additionally enforce smooth transitions from intentional to unintentional action during the third stage by employing a conditional random field.
While simulating the above-mentioned inherent motion biases cannot possibly cover the entire diversity of unintentional actions, it results in a powerful representation leveraged by the third stage of our approach and achieves a new state of the art for unintentional action recognition tasks.

\begin{figure*}[t!]
\centering
\includegraphics[width=0.99\linewidth]{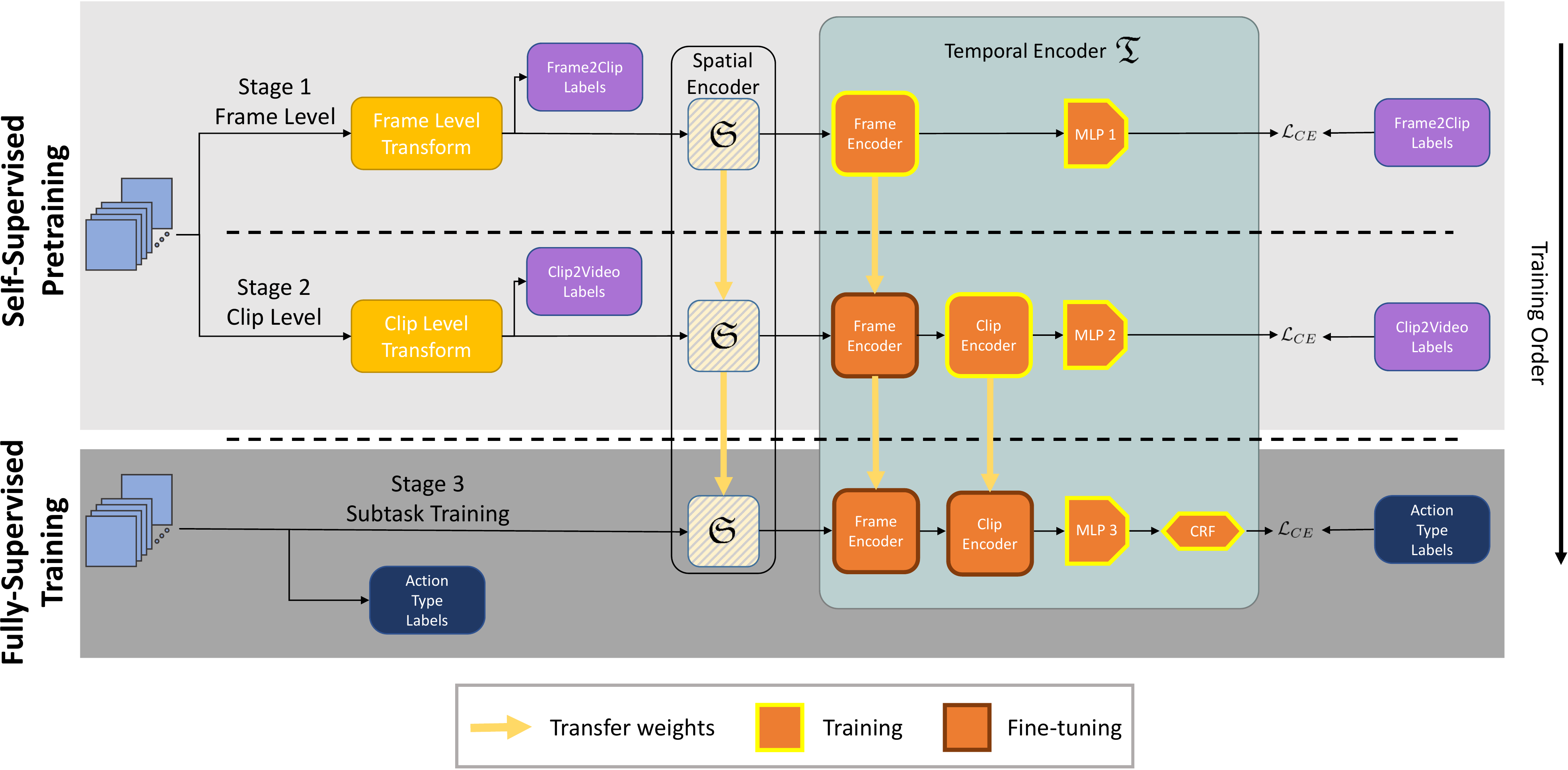}
\caption{\fontsizenine{Framework overview. In the first and the second stages, we use self-supervised \review{feature enhancement} by predicting \tib.
In the \textbf{first stage}, we \review{enhance the} representations \review{so} that \review{they} encode short-term dependencies based on neighbouring frames by training the frame encoder $\mathfrak{T}_{frame}$ and $MLP_1$. During the \textbf{second stage}, we \review{further fine-tune} representations \review{so} that \review{they} encode long-term dependencies using inter clip information by fine-tuning
the frame encoder $\mathfrak{T}_{frame}$ and training the clip encoder $\mathfrak{T}_{clip}$ together with $MLP_2$. During the \textbf{stage three}, we train in a fully-supervised way for downstream tasks by fine-tuning the frame $\mathfrak{T}_{frame}$ and clip $\mathfrak{T}_{clip}$ encoders, while we train $MLP_3$ and the CRF parameters.}}
\label{fig:overview}
\end{figure*}

For the first and the second stages, 
we generate labels based on the selected transformations. 
We extract frame features with a fixed pretrained spatial encoder, in particular using a pretrained ViT~\cite{dosovitskiy2020vit} model. \review{Furthermore, for fair comparison to previous work we also explore random initialization without any additional pre-training employing a Resnet3D architecture.} 
Then we pass the frame features to the temporal encoder. The architecture of the temporal encoder changes from stage to stage that we discuss in detail in Sec.~\ref{sec:stages}. For the first and the second stages, we use cross entropy loss with self-generated labels. The overview of the framework is shown in  Fig.~\ref{fig:overview}. 

\subsection{Temporal Transformations of Inherent Biases of Unintentional Actions (\tib)}
\label{subsec:trans}
\begin{figure}[t!]
\centering
\begin{minipage}{.48\textwidth}
  \centering
  \includegraphics[width=.8\linewidth]{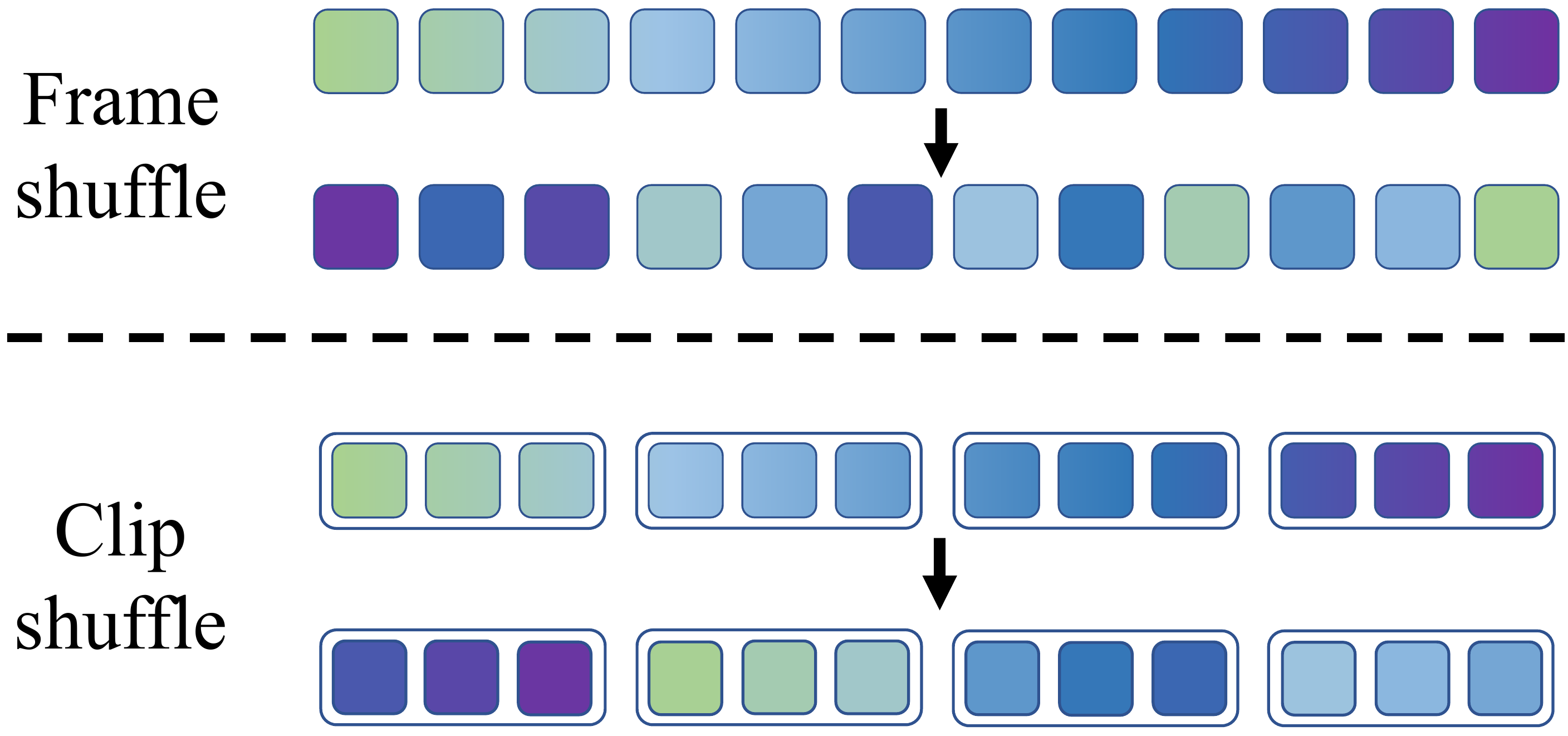}
  \captionof{figure}{\fontsizenine{Shuffle transformation. \textbf{Top}: Frame level. Each frame moves randomly to any position. \textbf{Bottom}: Clip level. We group frames into clips. Frames cannot change the position within a clip, while clips are shuffled randomly.}}
  \label{fig:frame_trc_2_clip_trn}
\end{minipage}%
\hfill
\begin{minipage}{.48\textwidth}
  \centering
  \includegraphics[width=.78\linewidth]{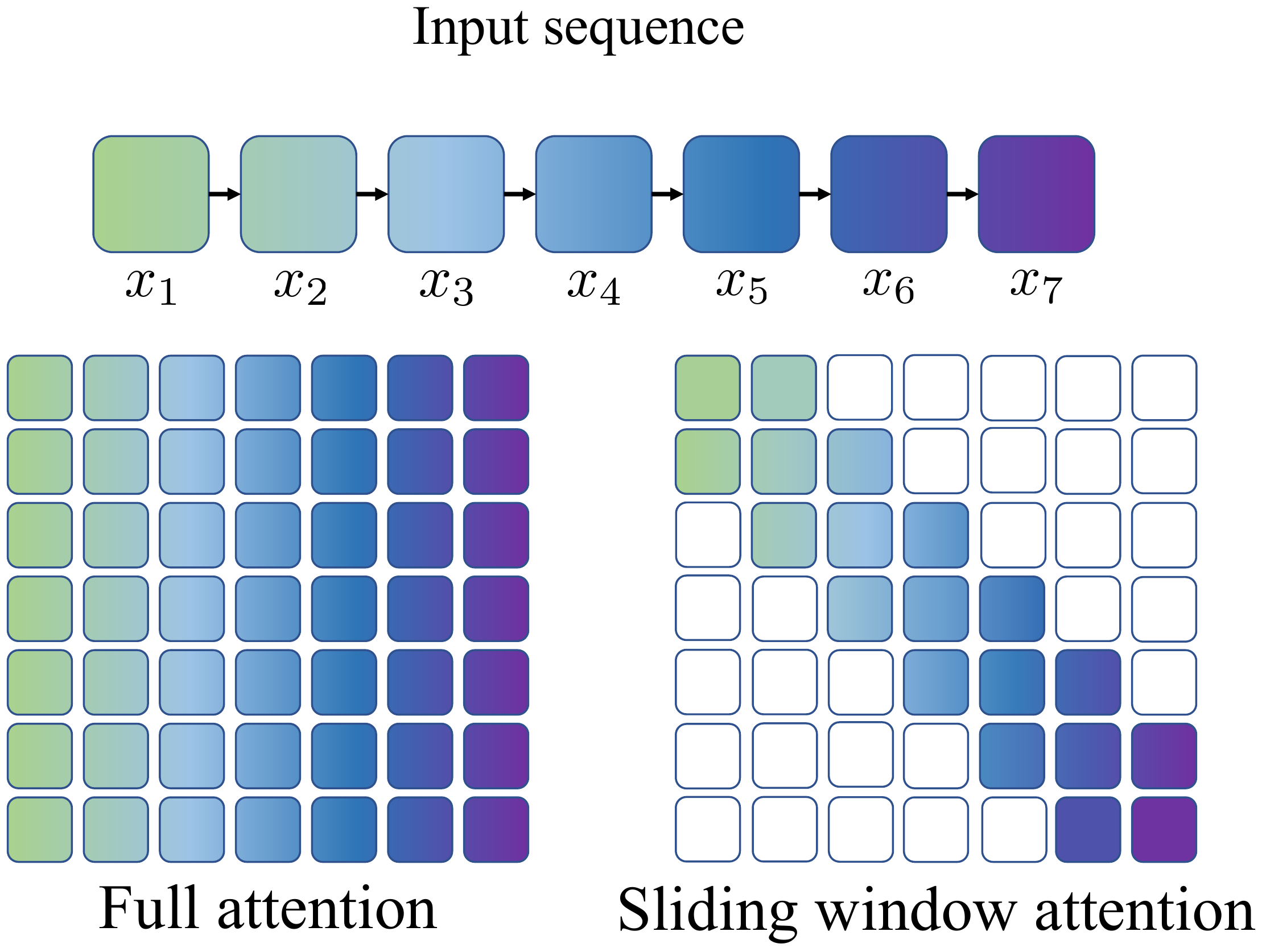}
  \captionof{figure}{\fontsizenine{Attention computation. \textbf{Left}: Full attention.  Required memory and processing time scale quadratically. \textbf{Right}: Sliding window attention mechanism. Required memory and processing time scale linearly with the input sequence length.}}
  \label{fig:longformer_attn}
\end{minipage}
\end{figure}

The potential diversity of unintentional actions encompasses all possible types of activities, since it is human nature to have failures during executing intentional actions.
We aim to grasp the motion inherent biases of unintentional actions and 
propose several temporal transformations grouped in two categories: motion speed and motion direction transformations.
Changes in the motion speed can correspond to  unintentional actions when, for example, the person stumbles and tries to keep balance by moving faster. Whereas, changes in the motion direction can occur by falling down or unexpectedly sliding backwards. While these are just two examples, in practice similar observations hold for a wide variety of unintentional actions.
We ground the set of transformations on the above intuition that connect temporal motion and inherent biases of unintentionality.
In this work, we consider each video as an ordered set of units, which can be either frames or clips. We formulate the framework in a multi-stage manner and, thus, apply these transformations in the first stage to frames and in the second stage to clips. In Fig.~\ref{fig:frame_trc_2_clip_trn} we show the difference between frame-level and clip-level \tib using the shuffle transformation.

\begin{figure}[ht!]
\centering
\includegraphics[width=0.5\linewidth]{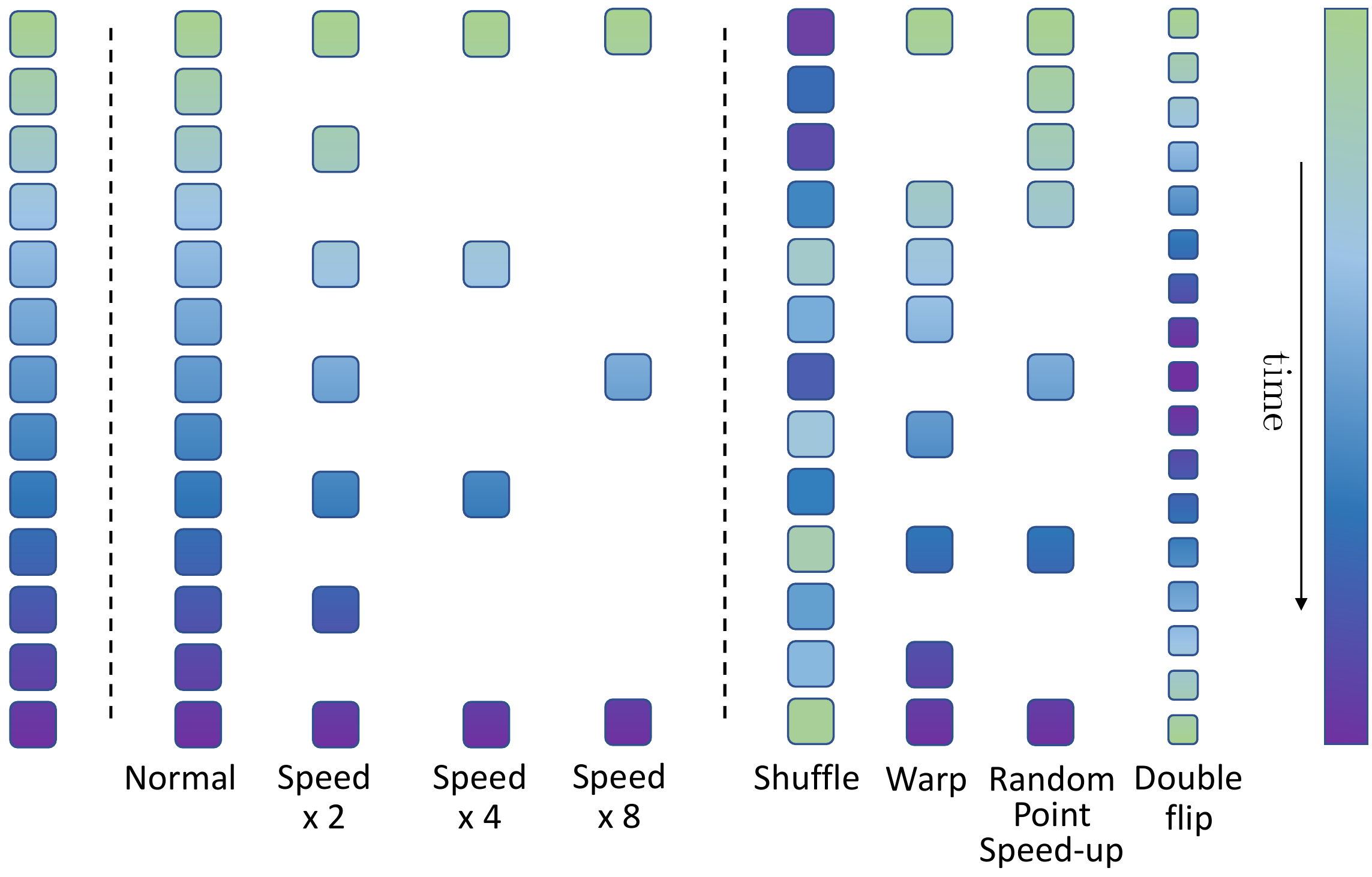}
\caption{ \fontsizenine{We group \tib in the \textbf{speed} group, where the speed of the video changes uniformly, and the \textbf{direction} group, where the speed changes non-uniformly, or we permute the units of the video.}}
\label{fig:temp_transformations}
\end{figure}

We define \tib as follows:
\begin{itemize}[itemsep=-0.1em]
    \item \emph{Speed-up}: We synthetically vary the number of units of the video by uniformly subsampling them with the ratios $\{\nicefrac{1}{2}, \nicefrac{1}{4}, \nicefrac{1}{8}\}$
    \item \emph{Random point speed-up}: We sample a random index $ri \in \{1 \cdots t\}$ to indicate the unit from which we are synthetically speeding up the video. Specifically, we start subsampling of the video units after $ri$ unit with ratio $\nicefrac{1}{\rho}$:
    \begin{equation*}
        [1, 2, \cdots, t-1, t ] \rightarrow [1,2,\cdots, ri, ri+\rho, \cdots, t-\rho, t].
    \end{equation*}
    where $t$ is the length of the video. 
    
    \item \emph{Double flip}: We mirror the sequence of the video units and concatenate them to the original counterpart:
    \begin{equation*}
        [1, 2, \cdots, t-1, t ] \rightarrow [1, 2, \cdots, t-1, t, t-1, \cdots, 2, 1 ].
    \end{equation*}
    
    \item \emph{Shuffle}: The video units are sampled in a random, non-repeating and non-sorted manner. 
    \item \emph{Warp}: The video units are sampled randomly and sorted increasingly by their original index.
    
\end{itemize}
All  transformations are depicted in Fig.~\ref{fig:temp_transformations}. We group \tib in the motion speed group where we include the \emph{Speed-up} variations and the motion direction group where we include the rest of the proposed \tibb.

\myparagraph{\tib prediction:} For 
\review{the feature enhancement}, we associate a discrete label to each \tib and train our framework in a self-supervised way to predict the label of the \tib 
that we apply either on frame or clip level. Each input sequence is transformed into a set of 6 sequences.
The correspondences between the self-generated labels and the 6 transformed sequences are as follows:
1 - Initial sequence without \tibb; \{2, 3, 4\} - \textit{Speed-up} with sampled ratio $\in \{\nicefrac{1}{2}, \nicefrac{1}{4}, \nicefrac{1}{8}\}$; 5 - \textit{Random point speed-up}; 6 - \textit{Double flip}; 7 - \textit{Shuffle}; 8 - \textit{Warp}.
Specifically, we predict the transformation applied to each of the 6 resulting sequences correspondingly.
Therefore, the model learns to distinguish various simulated inherent biases of unintentional actions on the same sequence.

\section{Multi-Stage Learning for Unintentional Action Recognition}
\label{sec:stages}
In this section, we first introduce the Transformer block, a building element for the different modules of our temporal encoder in Sec.~\ref{subsec:transformer}. Then we discuss in details each learning stage of our framework in Sec.~\ref{subsec:stage1},~\ref{subsec:stage2} and~\ref{subsec:stage3}. 
 
\subsection{Transformer block}
\label{subsec:transformer}
In this work, to process long sequences, we utilize a transformer architecture. We integrate intra and inter clip information and capture information for unintentional action recognition from different granularity levels, frame and clip levels.
To aggregate information over the whole video, we need to process a long sequence of consecutive frames or clips from the video. Further, we refer to the elements of frame or clip sequences as units. In the datasets, all videos are usually of different lengths, and the vanilla transformer model~\cite{dosovitskiy2020vit} scales quadratically with the sequence length in the required memory and processing time. 
The Longformer~\cite{beltagy2020longformer} addresses this problem by attending only to a fixed-sized window for each unit. The difference in the attention computation is shown in Fig.~\ref{fig:longformer_attn}.
We apply a window of size $w$ so that each unit attends to $\frac{1}{2}w$ units from the past and $\frac{1}{2}w$ from the future. 
All building blocks for the temporal encoder in our work follow the structure of  Longformer attention blocks. The transformer blocks can update the input sequence based on the temporal context or aggregate the temporal information from the entire sequence into one representation vector. For the former, for a given input sequence of units $U=\{u_1, \cdots, u_K\}$ we update the sequence and output $U^{\prime}=\{u_1^{\prime}, \cdots, u_K^{\prime}\}$. For the latter, we expand the given input sequence with an additional classification unit $cls$ that serves as aggregation unit across the whole sequence~\cite{dosovitskiy2020vit},~\cite{neimark2021vtn}. Therefore, we map the expanded input sequence $U \cup \{cls\}$ to $cls^{\prime}$ that aggregates the sequence into one representation. 
For more details we refer to the original paper~\cite{beltagy2020longformer}.


\subsection{[Stage 1] Frame2Clip (F2C) learning}
\label{subsec:stage1}
In the first stage of our framework, we enhance the pre-trained features such they encode intra clip information based on neighboring frames. We operate within short clips and, thus, capture local motion information. First, we sample a short clip $x$ of the length $n$ from a long input video $X$, where $n<|X|$ and $|X|$ is the length of the input video.
The clip $x$ consists of sequential frames $x=\{f_1, \cdots, f_n\}$ from the same video $X$.  We apply \tib $\mathcal{T}$ to this sequence of frames of the clip $x$ as we show schematically in the shuffle transformation on the top of Fig.~\ref{fig:frame_trc_2_clip_trn}. As a label for the transformed clip $\mathcal{T}(x)$ we use the index that corresponds to the respective \tibb, e.g. the shuffle transformation has index $7$. 
Then, from the spatial encoder $\mathfrak{S}$ we obtain the spatial frame features for the sequence.
Thereafter, to impose the intra connections between the frames of the same sequence, we use the temporal frame encoder $\mathfrak{T}_{frame}$ that is composed of the transformer blocks. The output of this encoder is one representation vector for the clip, hence we expand the sequence of frame features with an additional unit $cls_1$ to aggregate the temporal information from the clip $x$ into one representation $\tilde{x}$. Finally, we predict the \tib index with a classification layer $MLP_1$. We compute cross-entropy loss and optimize parameters of the temporal frame encoder $\mathfrak{T}_{frame}$ and a classification layer $MLP_1$.
Formally, we apply the following pipeline in the first stage:
\small
\begin{align}
    & \hat{y}_{clip} = MLP_{1}(\mathfrak{T}_{frame}(\mathfrak{S}(\mathcal{T}(f_1, \cdots, f_n)), cls_1)),
\end{align}
\normalsize 
where $\hat{y}_{clip}$ is the predicted label of the transformation applied to the clip $x$. Note that at this stage of the training, we process all sampled clips independently of each other. \review{Our F2C learning implies intra-clip encoding, therefore, it can be also substituted with the more common spatio-temporal networks such as ResNet3D that learn representations of short clips. }

\subsection{[Stage 2] Frame2Clip2Video (F2C2V) learning.}
\label{subsec:stage2}
In the second stage of our framework, we \review{further enhance the representations} based on the clip level transformations. At this stage, we integrate inter clip information into the video representation to model long-term dependencies. 
The input video $X$ of length $|X|$ we split into overlapping clips $z=\{x_1, x_2, \cdots, x_N\}$ that we sample with stride $k$ from the video, where $N=\frac{|X|-n}{k}+1$. Each clip consists of $n$ frames $x_i=\{f^i_1, \cdots, f^i_n\}$. During F2C2V learning stage we apply \tib $\mathcal{T}$ to the sequence of clips of the whole video $X$ as we show schematically on the bottom of Fig.~\ref{fig:frame_trc_2_clip_trn}. Specifically, the order of consecutive frames within each clip remains fixed, whereas the sequence of the  clips is transformed $\mathcal{T}(z)$. As in the first stage, we first pass frames of the clips through the spatial encoder $\mathfrak{S}$ to obtain frame-level features, then the sequence of frames for each clip $x_i=\{f^i_1, \cdots, f^i_n\}$ is aggregated into one clip representation $\tilde{x}_i$ by the frame encoder $\mathfrak{T}_{frame}$. Note that the frame sequence during this stage follows the original order. Then we aggregate the transformed sequence of clip representations $\mathcal{T}(\tilde{x}_1, \tilde{x}_2, \cdots, \tilde{x}_N)$ into video representation vector $\tilde{X}$ with the temporal clip encoder $\mathfrak{T}_{clip}$. We predict \tib that was applied to the sequence of clips with the classification layer $MLP_2$. Formally, the second stage is as follows:
\small
\begin{align}
    & \tilde{x}_i = \,\,\mathfrak{T}_{frame}(\mathfrak{S}(f_1^i, \cdots, f_n^i), cls_1) , \,\, \forall i \in \{1, \dots, N\}; \label{eq:clip_level}\\
    & \hat{y}_{video} = \,\,MLP_2(\mathfrak{T}_{clip}(\mathcal{T}(\tilde{x}_1, \cdots, \tilde{x}_N), cls_2)), \label{eq:video_level}
\end{align}
\normalsize
where $\hat{y}_{video}$ is the predicted \tib label applied to a sequence of clips from video $X$. To aggregate all the clips into one video vector we similarly use an additional classification token $cls_2$ as in the previous stage. At this stage, we optimize the parameters of the temporal clip encoder $\mathfrak{T}_{clip}$ and the classification layer $MLP_2$, the parameters of the temporal frame encoder $\mathfrak{T}_{frame}$ we transfer from the previous stage and fine-tune. Note that at each stage we utilize a new classification layer and discard the one from the previous stage.

\subsection{[Stage 3] Downstream Transfer to Unintentional Action Tasks}
\label{subsec:stage3}
In the last stage, we extend our framework to supervised unintentional action recognition tasks. For these tasks, we predict the unintentionality for each short clip rather than for the entire video. This stage is completely supervised with clip level labels, therefore we do not apply \tibb. The input video $X$ is divided into temporally ordered overlapping clips $z=\{x_1, x_2, \cdots, x_N\}$. We follow the pipeline of the second stage and extract clip level representation vectors $\{\tilde{x}_1, \tilde{x}_2, \cdots, \tilde{x}_N\}$ with the temporal frame encoder $\mathfrak{T}_{frame}$. Then we use the temporal clip encoder $\mathfrak{T}_{clip}$ without the classification aggregation unit $cls_2$ given that for the supervised tasks we use clip-level classification labels. The output of $\mathfrak{T}_{clip}$ then is the sequence of clip representations instead of one video representation vector. By using the temporal clip encoder $\mathfrak{T}_{clip}$ each clip representation vector is updated with the inter clip information from the whole video $X$. Finally, we pass the clip vectors through the classification layer $MLP_3$ to obtain the emission scores that we use in the CRF layer $\mathfrak{T}_{CRF}$. The overall downstream transfer stage is as follows:
\small
\begin{align}
    & \tilde{x}_i = \,\,\mathfrak{T}_{frame}(\mathfrak{S}(f_{i,1}, \cdots, f_{i,n}), cls_1) , \,\, \forall i \in \{1, \dots, N\}; \label{eq:clip_level_crf}\\
    & \{\hat{y}_1,\cdots,\hat{y}_N\} = \,\,\mathfrak{T}_{CRF}(MLP_3(\mathfrak{T}_{clip}(\tilde{x}_1, \cdots, \tilde{x}_N)), T), \label{eq:crf}
\end{align}
\normalsize
where $\{\hat{y}_1,\cdots,\hat{y}_N\}$ predicted clip level labels and $T$ is the transition matrix of the CRF layer. 

The CRF layer aims to explicitly model the high-level temporal structure of the unintentional videos between the clips as shown in Fig.~\ref{fig:transition_frames_fig}. For videos containing unintentional actions the high-level order imposes a smooth transition from normal to transition to unintentional clips, therefore we can incorporate this prior information on the order of the clip labels.
\begin{figure}[t!]
\centering
\includegraphics[width=0.6\linewidth]{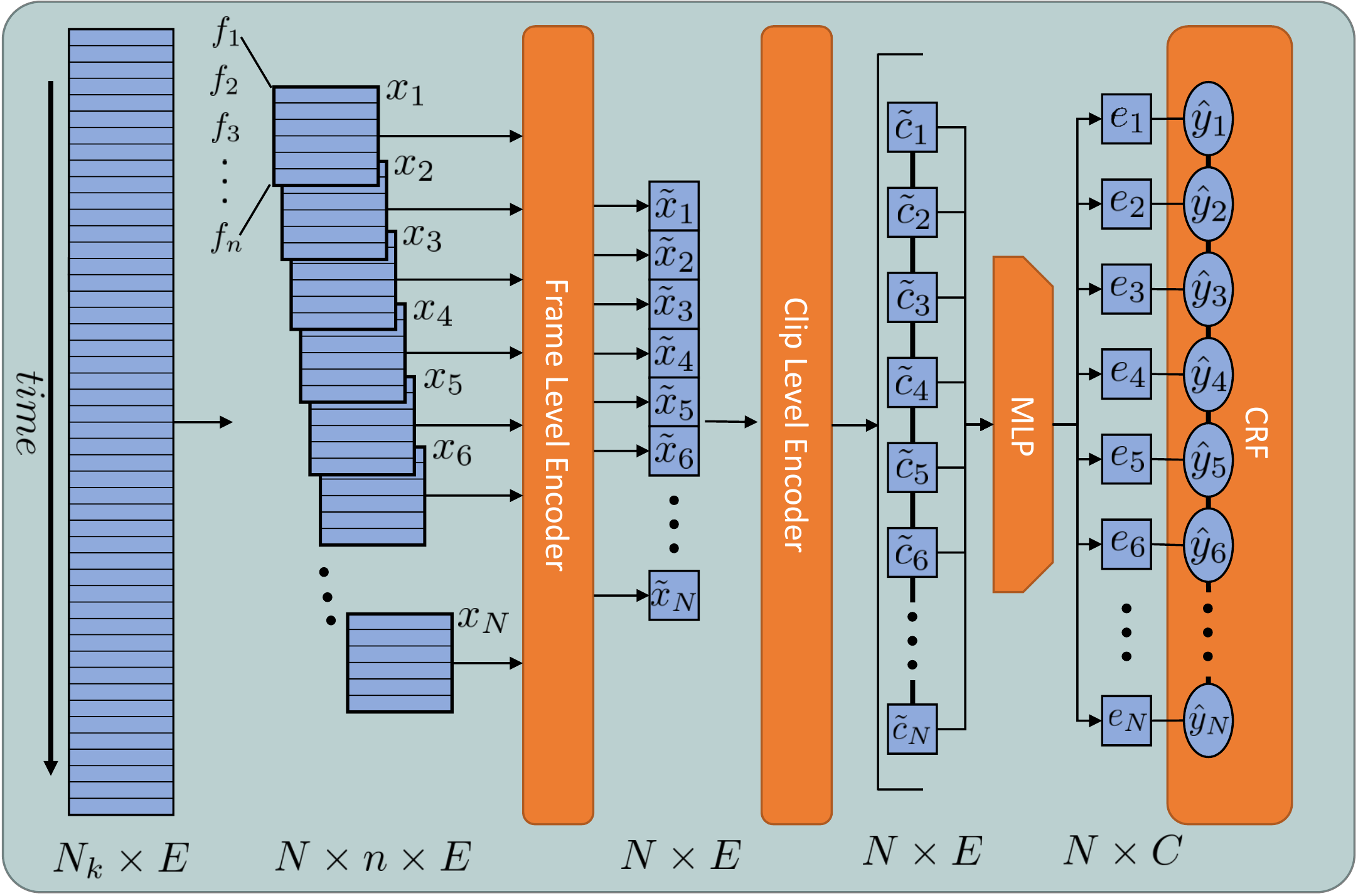}
\caption{\fontsizenine{Multi-stage temporal encoder. The $\mathfrak{T}_{frame}$ encoder aggregates the frames of each input clip $x_1$ into single-vector clip representation $\tilde{x}_i$. $\mathfrak{T}_{clip}$ updates $\tilde{x}_i$ with inter-clip information resulting into $\tilde{c}_i$. $MLP$ turns $\tilde{c}_1$ into emission scores $e_i$ for the CRF layer, which predicts clip-level labels.}} 
\label{fig:temp_encoder}
\end{figure}
The CRF layer facilitates dependencies between the current clip and the previously predicted labels and enforces neighbouring clips to have the same label.
This layer is parametrized by the transition matrix $T \in \mathcal{R}^{C \times C}$, where $T_{i, j}$ is the score of transitioning from label $i$ to label $j$ and $C$ stands for the number of classes. 
We denote by $\theta$ the parameters of the temporal encoder $\mathfrak{T}$ and $MLP_3$ together that we optimize during training, by $L\in \mathcal{R}^N$ the vector of clip labels for the input sequence of clips $z$ and by $E\in \mathcal{R}^{N\times C}$ an emission matrix that consists of logits for each unit of the input sequence generated by $MLP_3$ layer. 
Then, we can calculate the scores for a label sequence $\hat{L}$ given the input sequence $z=\{x_1, \cdots, x_N\}$, trainable weights $\theta$ and the transition matrix $T$ as
\small
\begin{equation}
    s(\hat{L}, z, \theta, T) = \sum_{t=1}^{n}\big(T_{L_{t-1}, L{t}} + E_{t, L_{t}}\big).
\end{equation}
\normalsize
Let $L_{all}$ be the set of all possible vectors of labels for the input clip sequence. Then the loss function is the negative log-likelihood:
\small
\begin{equation}
    \mathcal{L}(\theta, T) = -log\frac{exp(s(L_{gt}, z, \theta, T))}{\sum_{L\in L_{all}}exp(s(L, z, \theta, T))},
\end{equation}
\normalsize
where $L_{gt}$ the ground truth labels for each clip of the input sequence $z$. 
The total loss is the average over all training sequences. 

During the inference, we compute the optimal labels $L_{opt}$ for the input sequence of clips $z_{test}$ as
\small
\begin{equation}
    L_{opt} = \argmax_{L\in L_{all}} s(L, z_{test}, \theta^{\star}, T^{\star}), \label{eq:viterbi}
\end{equation}
\normalsize
where $\theta^{\star}$ and $T^{\star}$ are the optimized model parameters.  Note that Eq.~\ref{eq:viterbi} can be solved efficiently with the Viterbi algorithm.


%% file: 4_exp_results.tex
\section{Experimental Results}
\label{sec:exp_res}

In this section, we present experimental findings that validate our framework.
First, we introduce the Oops! dataset, \review{present different backbone models used in the framework} and discuss the implementation details of our method. We further compare our framework to the state-of-the-art on the three subtasks in Sec.~\ref{sec:sota_comparison} and then we present extensive ablation experiments to validate the impact of each of the components in Sec.~\ref{sec:ablation}.  

\myparagraph{Dataset:} Oops!~\cite{epstein2019unint_actions} is a collection of 20,338 amateur fail videos from YouTube. The dataset is split into three sets: 7,368 labelled videos for supervised training, 6,739 labelled videos for validation, and 6,231 unlabelled videos for pretraining. Each label in the first two sets consists of the timestamp where the action transitions from intentional to unintentional. Following the clip sampling procedure described in Sec.~\ref{sec:stages} we get $18,069$ intentional clips, $4,137$ transitional clips and $19,679$ unintentional clips. 

\myparagraph{Spatial Encoder:} 
\review{To disentangle the influence of different components of the framework, to provide a fair comparison to previous methods and to follow the trend of performant architectures, we use two backbones for the main experiments. Similarly to prior work~\cite{epstein2019unint_actions},~\cite{han2020mem_dpc}, we employ ResNet3D (R3D)~\cite{hara2017resnet3d}. This backbone is spatio-temporal and, therefore, substitutes both spatial $\mathfrak{S}$ and frame level $\mathfrak{T}_{frame}$ encoder as we discuss in Sec.~\ref{subsec:stage1}, we refer to it as F2C level for R3D backbone. In this setup, we learn the representations from scratch instead of enhancing pre-trained representations to fairly compare to previous works. To decouple spatial and temporal dimensions we use ViT model~\cite{dosovitskiy2020vit} pretrained on ImageNet-21K\cite{ridniki2021_21k} (IN-21K). In this setup, we leverage the pre-trained image representations and further enhance them for unintentional actions.
We note that we freeze ViT model, while R3D model we train from scratch with a random initialization. Each encoder consists of three stacked transformer blocks, each block constitutes of 16 parallel heads. Additional backbones are in the supplement. } 


\myparagraph{Implementation details:}
We sample clips of 16 frames from the input video with stride $k=4$.
We train for 100 epochs for the first and second stage and for 50 epochs for the third stage with the AdamW~\cite{loshchilov2017adamw} optimizer with weight decay of $1e-4$. The starting learning rate for all stages is $1e-4$ and is decreased to $1e-6$ using a cosine decay policy. During the third stage, the loss function is weighted as $\omega_i = \nicefrac{max(\eta_1, \eta_2, \cdots, \eta_{c})}{\eta_i}$,
where $\eta_1$ is the number of labels in class $i$ and $c$ is the number of classes. We recalculate the weights during the anticipation task as class sample distribution changes over time. More details can be found in the supplement.

\subsection{Comparison to state-of-the-art}
\label{sec:sota_comparison}
In this section we compare the performance of our framework on the three downstream tasks as classification, localization and anticipation. In Tab.~\ref{table:sota} we provide a comparison to the previous state-of-the-art method across all three tasks. 
The first row in the table corresponds to supervised pretraining on the Kinetics 700~\cite{smaira2020kin700}. The benchmark that we follow for the fair comparison across all the tasks is defined in Oops! dataset~\cite{epstein2019unint_actions}.

\input{tables/uar_sota_comparison_classification}

\myparagraph{Unintentional Action Classification.} 
We first compare our framework to the recent state-of-the-art work on unintentional action classification. \review{We divide Tab.~\ref{table:sota} into blocks with comparable backbones and pre-training methods. }
We can observe a significant improvement over the previous state-of-the-art method by \review{approximately} $10$ points \review{with R3D backbone and training from scratch}. In contrast to Han et al.~\cite{han2020mem_dpc} we pretrain our model only on the Oops! dataset without additional external data.  With our temporal encoder on the frame level, we gain minor improvement by \review{$0.9$ points indicating that \tib helps to improve even on the F2C level}, while our two stage temporal encoder allows us to achieve substantial increase in the performance by \review{$9.6$} points. \review{We notice further increase in performance when we use pre-trained frozen ViT as our backbone. For this setup, we enhance the pre-trained features and gain an improvement of $1.1$ points with our frame level encoder and $12.5$ points with our two stage temporal encoder.} These results indicate the importance of connections between intra and inter clip information in the video

\myparagraph{Unintentional Action Localisation.} For this downstream task, we localize the transition point from intentional to unintentional action in the video. We directly validate the network trained on unintentional action classification task and detect the transition point from intentional action to unintentional. The transitional clip for this task we define as the clip with the maximum output score of being transitional. In Tab.~\ref{table:sota} the localisation column shows the performance of our framework for two temporal localisation thresholds. \review{By using R3D as our backbone}, we improve by $2.8$ and $4.2$ points for the $\tau_L=0.25$ and $\tau_L=1$ respectively compared to Epstein et al.~\cite{epstein2019unint_actions}.\review{We imporove by $6.2$ and $7.5$ points for each threshold respectively when using ViT as our backbone.}  Additionally, we can observe the same trend as for the classification task, specifically, the clip level model outperforms the frame level model.  Note that for this task we reuse the model that is trained for the classification task where the number of normal and unintentional clips is notably higher than the number of transitional clips. The localisation task requires the model to be very specific on the boundaries when the unintentionality starts. We suppose that this influences the smaller improvement for the localization task than for the other tasks.

\myparagraph{Unintentional Action Anticipation.}
Further, we validate our framework on anticipation of unintentional actions. During the supervised training, we train our model to predict the label of the future clip. To directly compare to the previous work, we anticipate an action $1.5$ seconds into the future as in previous work.
Our model achieves new state-of-the-art results with a considerable improvement by $19.4$ \review{ and $21.4$ points for the R3D and ViT backbones respectively}. Note that the performance of our framework is better for the anticipation task than for classification. As mentioned, the dataset includes more unintentional clips than normal clips, and in combination that we are able to predict unintentional clips more accurately, it leads to the difference in the performance between the downstream tasks. 

\subsection{Ablation study}
\label{subsec:ablation_study}
We perform ablation studies on different components and backbones of our framework to assess their influence on the overall performance. We use the unintentional action classification task for all the evaluations in this section. \review{Extended ablations and qualitative results can be found in the supplement.}

\myparagraph{Feature enhancement stages.} In this section we analyse the influence of the self-supervised \review{training to enhance the representations} with \tib on the UA classification task. In Tab.~\ref{table:uar_ablation_crf_influence} the first and the second rows show the performance with and without \tib respectively. Specifically, for the first row, we skip the self-supervised procedure for the pretraining of the parameters and directly optimize from scratch the respective temporal encoders with the supervised task. For the second row, we include the self-supervised feature enhancement corresponding to the respective learning stages.
We can observe that for each temporal encoder (F2C and F2C2V) we have a significant increase with \tib representation \review{enhancing (learning)}. Considering only intra clip information (F2C) we improve by $4.6$ \review{and $4.5$ points for the ViT and R3D backbones respectively} while using additionally inter clip information we improve by $4.7$ \review{and $4.8$} points. These results confirm the importance of the self-supervised feature enhancement for UA recognition performance.

\begin{table}[t!]
\centering
\begin{minipage}{.58\textwidth}
  \centering
    \caption{\fontsizenine{Ablation results: Influence of \tib and CRF layer on classification performance for two representations, pretrained ViT representation and R3D.}}
    \label{table:uar_ablation_crf_influence}
    \begin{tabular}{c|c|cc|cc}
    \toprule
    \multirow{2}{*}{\tib} & \multirow{2}{*}{CRF} & \multicolumn{2}{c}{ViT}  & \multicolumn{2}{c}{R3D} \\
    & & F2C & F2C2V & F2C & F2C2V \\ 
    \midrule
    $-$ & $-$ &  60.9 & 69.6 & 59.3 & 65.6\\
    \ding{52}&$-$ &  65.5 & 74.3 & 63.8 & 70.4 \\
    \ding{52}&\ding{52} & 65.0 & \textbf{76.9} & 65.3 & \textbf{74.0}\\
    \bottomrule
    \end{tabular}
\end{minipage}%
\hfill
\begin{minipage}{.4\textwidth}
  \centering
   \caption{Influence of \tib groups on UA classification task for different stages of learning of the temporal encoder.}
    \label{table:uar_ablation_trn_type}
  \begin{tabular}{c|cc}
    \toprule
    \tib & F2C & F2C2V \\
    \midrule
    $-$& 60.9  & 73.2\\
    Speed & 62.5  & 73.8\\
    Direction & 63.3  & 75.1\\ %
    All & \textbf{65.5} & \textbf{76.9}\\
    \bottomrule
    \end{tabular}
\end{minipage}
\end{table}

\myparagraph{Learning Stages for Temporal Encoder.} 
In this section, we evaluate the impact of multi-stage learning of our temporal encoder. In Tab.~\ref{table:uar_ablation_crf_influence} we show in the F2C column the performance across different settings for the corresponding encoder, see the detailed structure in Fig.~\ref{fig:temp_encoder} \review{for ViT backbone, whereas R3D backbone comprises this stage}. In the second column with the F2C2V encoder, we observe a significant improvement by about $5-10$ points consistently across all the settings. Furthermore, we evaluate the performance of the spatial features from the fixed pretrained ViT model~\cite{dosovitskiy2020vit} that we use as input features to our temporal encoder. We obtain $58.4$ points without \tib and without CRF. It supports the importance of the temporal encoder that is able to capture successfully inter and intra clip information.

\myparagraph{\tib groups.} We assess the influence of the \tib groups on the overall performance. Tab.~\ref{table:uar_ablation_trn_type} 
shows that both groups improve the performance for both encoder levels. We notice that the direction \tib have a more significant impact than the speed \tibb. Whereas, in contrast to the separate groups, the combination of the speed and the direction transformations leads to \review{greatly enhanced} representations that effectively capture the inherent biases of UA. \review{We additionally study an influence of each transformation separately that shows random point speed-up to be the most important, while the combination of all \tib is still predominant. More detailed study is in the supplement.}


\label{sec:ablation}


%% file: tables/uar_sota_comparison_classification.tex
\begin{table*}[t!]
\centering
\small
\tabcolsep=0.1cm
\caption{\fontsizenine{Comparison of our approach to state-of-the-art for UA classification, localisation and anticipation. F2C denotes the first (frame2clip) learning stage, F2C2V denotes the second (frame2clip2video) learning stage. $\tau_L$ indicates a time window in seconds for ground truth assignment for the localization task. $\tau_A$ indicates the time step in the future that we predict. 
Init. indicates fully supervised initialization of the backbone if applies. $^*$ indicates that the backbone is frozen during all stages.}}
\resizebox{\textwidth}{!}{
\begin{tabular}{c|ccc|cccc}
\toprule
\multirow{2}{*}{Method} & \multirow{2}{*}{Backbone} & \multirow{2}{*}{Init.} & Pretrain  & Cls. & \multicolumn{2}{c}{Loc.}&Ant.\\
 & & & Dataset & Acc. & $\tau_L:0.25$  & $\tau_L:1$ &  $\tau_A:1$\\
\midrule
K700 Supervision~\cite{epstein2019unint_actions} & R(18)3D & K700 &  - & 64.0 & 46.7 & 75.9 & 59.7 \\
\midrule
Epstein et al. \cite{epstein2019unint_actions} & R(18)3D & - & Oops! & 61.6  & 36.6 & 65.3 & 56.7\\
Han et al. \cite{han2020mem_dpc} & (2+3D)R18 & - & (K400+Oops!) & 64.4 & - & - & - \\
Ours (F2C) & R(18)3D & - & Oops! &  65.3 & 37.7 & 67.8 & 66.7\\
Ours (F2C2V) & R(18)3D & - & Oops! &  \textbf{74.0} & \textbf{39.4} & \textbf{69.5} & \textbf{76.1} \\
\midrule
Ours (F2C) & ViT$^*$ & IN-21K & Oops! & 65.5 & 41.4 & 72.2 & 69.2\\
Ours (F2C2V) & ViT$^*$ & IN-21K & Oops! & \textbf{76.9} & \textbf{42.8} & \textbf{72.8} & \textbf{78.1} \\
\bottomrule
\end{tabular}}
\label{table:sota}
\end{table*}

%% file: 5_conclusions.tex

\section{Conclusion}
In this paper, we propose a multi-stage framework to exploit inherent biases that exist in videos of unintentional actions. First, we simulate these inherent biases with the temporal transformations that we employ for self-supervised \review{training to enhance pre-trained representations.} We formulate the representation enhancement task in a multi-stage way to integrate inter and intra clip video information. This leads to powerful representations that significantly enhance the performance on the downstream UA tasks. Finally, we employ CRF to explicitly model global dependencies. We evaluate our model on the three unintentional action recognition tasks such as classification, localisation, and anticipation, and achieve state-of-the-art performance across all of them.

\label{sec:concluson}

%% file: tables/uar_ablation_backbone_finetune.tex
\begin{table}[]
    \centering
    \caption{Influence of backbone model and representation initialization method on the overall framework performance.}
    \begin{tabular}{cccc|cc}
    \toprule
    \multirow{2}{*}{Backbone} & Frozen & Init.  & Init.  & \multirow{2}{*}{F2C} & \multirow{2}{*}{F2C2V}  \\
    & Backbone &  Method &  Dataset &  &  \\
    \midrule
    R(18)3D & - & - & - & 65.3 & 74.0 \\
    R18 & \checkmark & FS & IN 1K & 60.3 & 74.4 \\
    R50 & \checkmark &  FS & IN 21K & 61.2 & 74.9 \\
    ViT & \checkmark &  FS & IN 21K & 65.5 & 76.9 \\
    \bottomrule
    \end{tabular}
    \label{tab:backbone_finetune}
\end{table}

%% file: tables/uar_trn_ablation.tex
\begin{table}[]
    \centering
    \caption{Influence of each temporal transformation on the overall framework performance.}
    \begin{tabular}{c|cc|cc}
    \toprule
        Transformation &\,\, F2C\,\, & drop \,\, & F2C2V & drop\\
        \midrule
        All & 65.5 & - & 76.9 & -\\
        \midrule
        Shuffle & 63.9 & -1.6 & 75.6 & -1.3 \\ 
        Warp & 64.8 & -0.7 & 76.2 & -0.7\\ 
        Random Point Speedup & 62.5 & \textbf{-3.0} & 74.7 & \textbf{-2.5} \\ 
        Double Flip & 63.5 & -2.0 & 76.1 & -0.8 \\ 
        Speed $\times2$ & 63.5 & -2.0 & 75.5 & -1.4\\ 
        Speed $\times4$ & 63.3 & -2.2 & 76.0 & -0.9\\ 
        Speed $\times8$ & 62.9 & -2.6 & 76.1 & -0.8\\ 
    \bottomrule
    \end{tabular}
    \label{tab:trn_ablation}
\end{table}

%% file: tables/uar_ablation_win_size.tex
\begin{table}[H]
\centering
\small
\tabcolsep=0.1cm
\caption{Performance on UA classification task with different sizes of the attention window.}
\label{table:uar_ablation_win_size}
\begin{tabular}{c?c|c|c|c}  
\toprule
$w$ & 4 & 16 & 32 & 64\\
\midrule
Accuracy & 63.8  & 64.3  & 65.5  & 64.7\\

\bottomrule
\end{tabular}
\end{table}

%% file: tables/uar_ablation_anticipation.tex
\begin{table}[H]
\centering
\small
\tabcolsep=0.1cm
\caption{Evaluation of the performance for different future time points on UA anticipation task. $\tau_A$ indicates the time step in the future that we predict.}
\label{table:uar_ablation_anticipation}
\begin{tabular}{c?c|c|c|c|c}  
\toprule
$\tau_A$ (sec) & 0 & 0.5 & 1 & 1.5 & 2\\
\midrule
Accuracy & 76.9  & 77.6  & 77.2  & 78.1 & 79.3\\

\bottomrule
\end{tabular}
\end{table}

%% file: tables/uar_ablation_lnf_layers.tex


\begin{table}[H]
 \caption{Influence of the number of transformer layers on the overall performance of the framework.}
    \centering
    \small
    \tabcolsep=0.1cm
    \begin{tabular}{c|cc}
    \toprule
    $\#$Layers & F2C & F2C2V \\
    \midrule
    1 & 60.9  & 73.7\\
    3 & \textbf{65.5}  & \textbf{74.3}\\
    6 & 64.8  & 71.2\\
    \bottomrule
    \end{tabular}
    \label{table:uar_ablation_lnf_layers}
\end{table}